\documentclass[final,1p,times]{elsarticle}

\usepackage{amssymb}
\usepackage{graphicx}%
\usepackage{multirow}%
\usepackage{amsmath,amssymb,amsfonts}%
\usepackage{amsthm}%
\usepackage{mathrsfs}%
\usepackage[title]{appendix}%
\usepackage{xcolor}%
\usepackage{textcomp}%
\usepackage{manyfoot}%
\usepackage{booktabs}%
\usepackage{algorithm}
\usepackage{algorithm}%
\usepackage{algorithmicx}%
\usepackage{algpseudocode}%
\usepackage{listings}%
\usepackage{float}%
\usepackage{tikz}%
\usepackage{subcaption}%
\usepackage{tabularx}  
\usepackage{multirow}  
\usepackage{lscape} 
\usepackage{longtable}
\usepackage{booktabs}
\usepackage{boldline}
\usepackage{siunitx}
\usepackage{xcolor}

\journal{}

\begin{document}

\begin{frontmatter}





\author[1]{Usman Gani Joy}\corref{cor1}\ead{usmanjoycse@gmail.com}

\affiliation[1]{
organization={School of Science, Engineering \& Technology, East Delta University}, 
addressline ={Abdullah Al Noman Road, Noman Society},
city={Chattagram},
postcode={4209},
country={Bangladesh}
 }



\title{Wavelet-Enhanced Neural ODE and Graph Attention for Interpretable Energy Forecasting}




\begin{abstract}
	Accurate forecasting of energy demand and supply is critical for optimizing sustainable energy systems, yet it is challenged by the variability of renewable sources and dynamic consumption patterns. This paper introduce a neural framework that integrates continuous-time Neural Ordinary Differential Equations (Neural ODEs), graph attention, multi-resolution wavelet transformations, and adaptive learning of frequencies to address the issues of time series prediction.The model employs a robust ODE solver, using the Runge-Kutta method, paired with graph-based attention and residual connections to better understand both structural and temporal patterns. Through wavelet-based feature extraction and adaptive frequency modulation, it adeptly captures and models diverse, multi-scale temporal dynamics. When evaluated across seven diverse datasets—ETTh1, ETTh2, ETTm1, ETTm2 (electricity transformer temperature), and Waste, Solar, and Hydro (renewable energy)—this architecture consistently outperforms state-of-the-art baselines in various forecasting metrics, proving its robustness in capturing complex temporal dependencies. Plus, the model enhances interpretability through SHAP analysis, making it perfect for sustainable energy applications.
\end{abstract}


\end{frontmatter}


\section{Introduction}\label{sec1}

The global transition to a sustainable energy future demands innovative technologies that optimize energy generation, storage, and utilization across pivotal sectors such as transportation, heating, and power systems. A crucial element in this change is the ability to accurately predict energy demand and supply—a feat made challenging by the introduction of renewable energy sources, changing consumption patterns, and weather-dependent environmental factors like weather changes. Time series prediction is a key aspect of this endeavor, facilitating key applications like renewable energy forecasting, grid system optimization, and energy-efficient building design. Energy data, typically gathered from heterogeneous sensor networks comprising solar photovoltaic (PV) panels, wind turbines, and smart meters, pose unique challenges on account of their high dimensionality, non-uniform sampling frequencies, and dynamic characteristics that evolve across various time horizons. The above characteristics are consequences of the variability of renewable energy generation, shifting consumption patterns, and exogenous circumstances like temperature fluctuations and wind patterns. To overcome the above challenges, novel information integration methods are needed that can synthesize diverse inputs of data—temporal variability, feature dependencies, and multi-scale patterns—into strong, scalable frameworks that facilitate strategic energy planning, system integration, and development of effective policies. Conventional forecasting techniques tend to struggle with satisfying different conditions of non-stationary character of contemporary energy systems and therefore are not able to provide an accuracy needed and flexibility required for sustainable applications. Such conditions emphasize the necessity for advanced, fusion-based solutions, as proposed by Chen et al. in examining fusion strategies for energy forecasting \cite{CHEN2023fusion}.

A key innovation in continuous-time modeling has been realized through the introduction of Neural Ordinary Differential Equations (Neural ODEs) by Chen et al., which reimagined neural networks via the parameterization of hidden state derivatives \cite{chen2018neural}. The innovation allows for outstanding flexibility for time series prediction, particularly in energy systems, where observations are frequently received aperiodically owing to random sensor sampling or missing observations—a phenomenon prevalent in solar and wind applications. Although promising, initial Neural ODE models were computationally inefficient and could not incorporate the multi-scale dependencies inherent in energy load data such as short-term demand peaks and long-term seasonal trends. Building on this, Rubanova et al. developed the Neural ODE framework further to more appropriately deal with irregularly sampled data to better accommodate the energy settings with different sensor frequencies \cite{rubanova2019latent}. Yet, these adaptations still had to contend with processing lengthy sequences and capturing in-depth feature interdependencies—both of which are vital for effective energy forecasting. Further development was provided with Kidger et al.'s Neural Controlled Differential Equations (Neural CDEs), which implemented continuous-time modeling to energy data, though stability and interpretability issues persisted, especially in multi-scale fusion tasks \cite{kidger2021neural}. Massaroli et al. investigated various stability issues; however, the effective incorporation of multi-scale temporal patterns within the non-stationary framework of energy data remained an issue \cite{massaroli2022stable}.

Concurrently, physics-informed approaches have enriched energy time series modeling by embedding domain knowledge into predictive frameworks, enhancing interpretability. Physics-informed neural networks (PINNs), for instance, blend physical principles with data-driven architectures, improving generalization and stability for tasks like load forecasting \cite{sholokhov2023physics}. While effective in controlled settings, PINNs struggle to incorporate multi-source, data-driven features—such as dynamic load patterns from diverse sensors—limiting their fusion capabilities across heterogeneous energy datasets. Meanwhile, contrastive learning techniques, exemplified by Woo et al.’s Seasonal-Trend (CoST) framework, disentangle seasonal and trend components to enhance multivariate forecasting accuracy \cite{woo2022}. However, CoST’s reliance on predefined seasonal patterns restricts its adaptability to the highly variable, non-stationary nature of energy data, where flexible fusion of adaptive features is critical for capturing real-world dynamics.

In building energy management—a key domain for sustainable energy technologies—the need for interpretable forecasting models has become increasingly urgent. Moon et al. proposed an explainable electrical load forecasting (XELF) methodology for educational buildings, evaluating five tree-based ensemble models: Random Forest (RF), Gradient Boosting Machine (GBM), XGBoost, LightGBM, and CatBoost \cite{moon2022sus}. Using Shapley Additive Explanations (SHAP) to interpret the top-performing LightGBM model, their work revealed the critical influence of historical electricity consumption and weather factors—such as the temperature-humidity index (THI) and wind chill temperature (WCT)—on forecasting accuracy. This transparency enhances its value for building energy management systems, where understanding prediction drivers is essential for decision-making. However, its focus on educational buildings limits its generalizability to other structures, like commercial or residential buildings with distinct usage patterns. Additionally, while tree-based models excel with structured data, they often fail to capture the complex temporal dependencies better addressed by time-series-specific architectures like recurrent neural networks (RNNs).

Deep learning has further advanced energy load forecasting, tackling specific challenges in sustainable energy systems. Luo et al. developed an adaptive Long Short-Term Memory (AD-LSTM) framework for day-ahead photovoltaic power generation (PVPG) forecasting, addressing concept drift from weather and system changes \cite{luo2022sus}. Through a two-phase adaptive learning approach, AD-LSTM refines an initial LSTM model with recent data, boosting accuracy and efficiency without full retraining. Though it outperforms standard LSTMs, its reliance on high-quality recent data, limited scalability for large systems, and reduced interpretability highlight areas for improvement. Similarly, Ma et al. explored the techno-economic feasibility of load forecasting for microgrid optimization, combining Particle Swarm Optimization (PSO), LSTM, and XGBoost for long-term hourly predictions \cite{ma2025sus}. Yet, its dependence on specific datasets and HOMER software assumptions, alongside computational complexity and tuning demands, may hinder real-time or broad deployment. Cobos-Maestre et al.’s Solar Wind Attention Network (SWAN) leverages multi-head attention for solar wind speed forecasting, achieving a one-day lead time MAE of 30.61 km/s \cite{cobos2024expert}. While outperforming models like WindNet for short-term forecasts, its autoregressive design limits adaptability to new coronal holes or mass ejections without atmospheric data, and its narrow focus restricts broader energy applications. Kim et al.’s TMF-GNN, a temporal matrix factorization-based graph neural network, excels in multivariate time series (MTS) forecasting with missing values, achieving an MAE of 194.889 on the GFT dataset \cite{kim2025expert}. However, its graph regularization falters with sparse variables, and its training complexity ($O(N^2D + ND^2 + CTK)$) and hyperparameter sensitivity challenge scalability. Gil-Gamboa et al.’s deep feed-forward neural network (DFFNN) for residential water consumption forecasting achieves a MAPE of 5.59\% over quarterly horizons \cite{gil2024expert}. Though preprocessing and clustering enhance its accuracy, its static design lacks flexibility for multi-scale dynamics or missing data prevalent in energy contexts.

Transformer-based architectures have also gained traction in energy load forecasting, leveraging their strength in modeling long-range dependencies—crucial for sustainable energy applications. Madhusudhanan et al.’s Yformer, inspired by U-Net transformers, fuses multi-scale features via skip connections, but its complexity and reconstruction losses hinder generalization \cite{kiran2022}. Saeed and Aldera’s PCA-adaptive transformer prioritizes key inputs for renewable energy forecasting, yet its computational cost and overfitting risk limit fusion efficiency \cite{Saeed_forecast}. Shen et al.’s Good Beginning Transformer (GBT) mitigates overfitting with a two-stage approach, though its rigid initialization restricts multi-source signal integration \cite{shen2023}. Wavelet-based methods, utilizing Daubechies wavelets, excel at localizing patterns in time and frequency domains, showing promise for non-stationary energy signals from multiple sensors \cite{wavelet2019}. Meanwhile, Ma et al.’s Functional Echo State Network (FESN) extends traditional Echo State Networks for time series classification, but its reliance on functional basis expansion constrains its ability to fuse high-dimensional, non-stationary energy data efficiently \cite{ma2016functional}.

Despite these advances, a persistent gap remains in developing models that seamlessly fuse multi-scale, multi-source information—spanning continuous dynamics, feature interdependencies, and localized patterns—while balancing adaptability, efficiency, and practical utility for sustainable energy forecasting. To bridge this gap, this research introduces a hybrid architecture tailored for information fusion in energy load forecasting. The proposed model integrates Neural ODEs, solved via the Runge-Kutta method, with graph attention mechanisms, adaptive frequency transforms, and wavelet-based analysis to address the multifaceted challenges of multi-sensor energy data. The proposed hybrid architecture for energy load forecasting leverages advanced neural network techniques, drawing inspiration from applications in diverse domains such as agricultural disease prediction \cite{reviewer_paper_1} and energy-efficient sensor network protocols \cite{reviewer_paper_2}, to integrate multi-scale, multi-source data for enhanced predictive accuracy and interpretability in sustainable energy systems. By combining continuous-time dynamics from Neural ODEs, relational insights via dynamic graph construction, and multi-scale feature extraction through parametric and Daubechies transforms, this approach offers a comprehensive solution for applications like renewable energy integration, grid optimization, and policy-driven energy assessments. By combining continuous-time dynamics from Neural ODEs, relational insights via dynamic graph construction, and multi-scale feature extraction through parametric and Daubechies transforms, this approach offers a comprehensive solution for applications like renewable energy integration, grid optimization, and policy-driven energy assessments. Enhanced computational efficiency, scalability, and multi-scale analysis are complemented by improved interpretability through SHAP analysis, providing actionable insights into feature contributions and decision-making—attributes vital for real-world sustainable energy systems.

\subsection*{Main Contributions}

\begin{itemize}
	\item \textbf{Fusion Architecture}: A novel integration of Neural ODEs (solved via the Runge-Kutta method), graph attention mechanisms, and wavelet transforms to process multi-source energy load data effectively.
	\item \textbf{Dynamic Graph-Based Fusion}: Enhanced multivariate time series analysis through dynamic graph construction, capturing inter-feature dependencies in energy systems.
	\item \textbf{Multi-Scale Feature Integration}: Incorporation of parametric frequency and Daubechies transforms to fuse periodic and localized patterns within non-stationary energy data.
	\item \textbf{Comprehensive Evaluation}: Rigorous testing across diverse energy-related datasets, showcasing superior fusion performance compared to state-of-the-art benchmarks.
	\item \textbf{Theoretical Analysis}: In-depth assessment of robustness, efficiency, and scalability, underscoring the model’s viability for sustainable energy applications.
	\item \textbf{Enhanced Interpretability}: Utilization of SHAP analysis to illuminate the influence of fused features on forecasting decisions, bolstering practical applicability.
\end{itemize}
	
\section{Multi-Path Neural ODE Fusion for Time Series Prediction}

This section provides a comprehensive description of the proposed model’s architecture and operational mechanisms, integrating a neural ordinary differential equation (Neural ODE) system with graph attention mechanisms and advanced feature extraction techniques. It is organized into subsections detailing the model’s overall structure, individual components, their complementary interactions, computational complexity, and algorithmic workflow.

\subsection{Model Architecture}

The model is designed as a multi-stage pipeline to process an input vector \( x \in \mathbb{R}^{d} \), where \( d \) represents the number of features in a multivariate time series. The pipeline commences with the construction of a dynamic adjacency matrix derived from feature correlations, which serves as a relational guide for the graph attention mechanism embedded within the Neural ODE component. This initial step establishes a data-driven foundation, allowing the model to focus on meaningful inter-feature relationships rather than assuming fixed connections.

Subsequently, the input \( x \) is channeled through four parallel processing paths, each meticulously crafted to extract a specific aspect of the time series data:
\begin{enumerate}
	\item Neural ODE Path with Graph Attention: This path projects the input into a higher-dimensional space (e.g., \( \mathbb{R}^{d \times h} \), where \( h \) is the hidden dimension), evolves it through a Neural ODE informed by graph attention, and applies global average pooling to produce a condensed representation capturing both temporal dynamics and feature interdependencies. The use of continuous-time modeling here ensures flexibility in handling irregular time steps.
	\item Daubechies Transform Path: Leveraging a learnable convolutional transform inspired by Daubechies wavelets, this path focuses on multi-scale temporal features, such as short-term fluctuations or localized transients, which are critical in datasets with abrupt changes. It outputs a feature vector that emphasizes these patterns.
	\item Parametric Transform Path: This path employs a learnable frequency-adaptive transform to identify and model periodic patterns inherent in the data, adapting to varying cycle lengths through trainable parameters. It is particularly effective for time series exhibiting rhythmic behaviors.
	\item Direct Dense Path: A straightforward dense layer processes the input to retain a direct, untransformed representation of the raw data, ensuring that essential baseline information is preserved alongside more complex transformations.
\end{enumerate}

\begin{figure}[ht]
	\centering
	\includegraphics[width=0.8\textwidth,height=9cm]{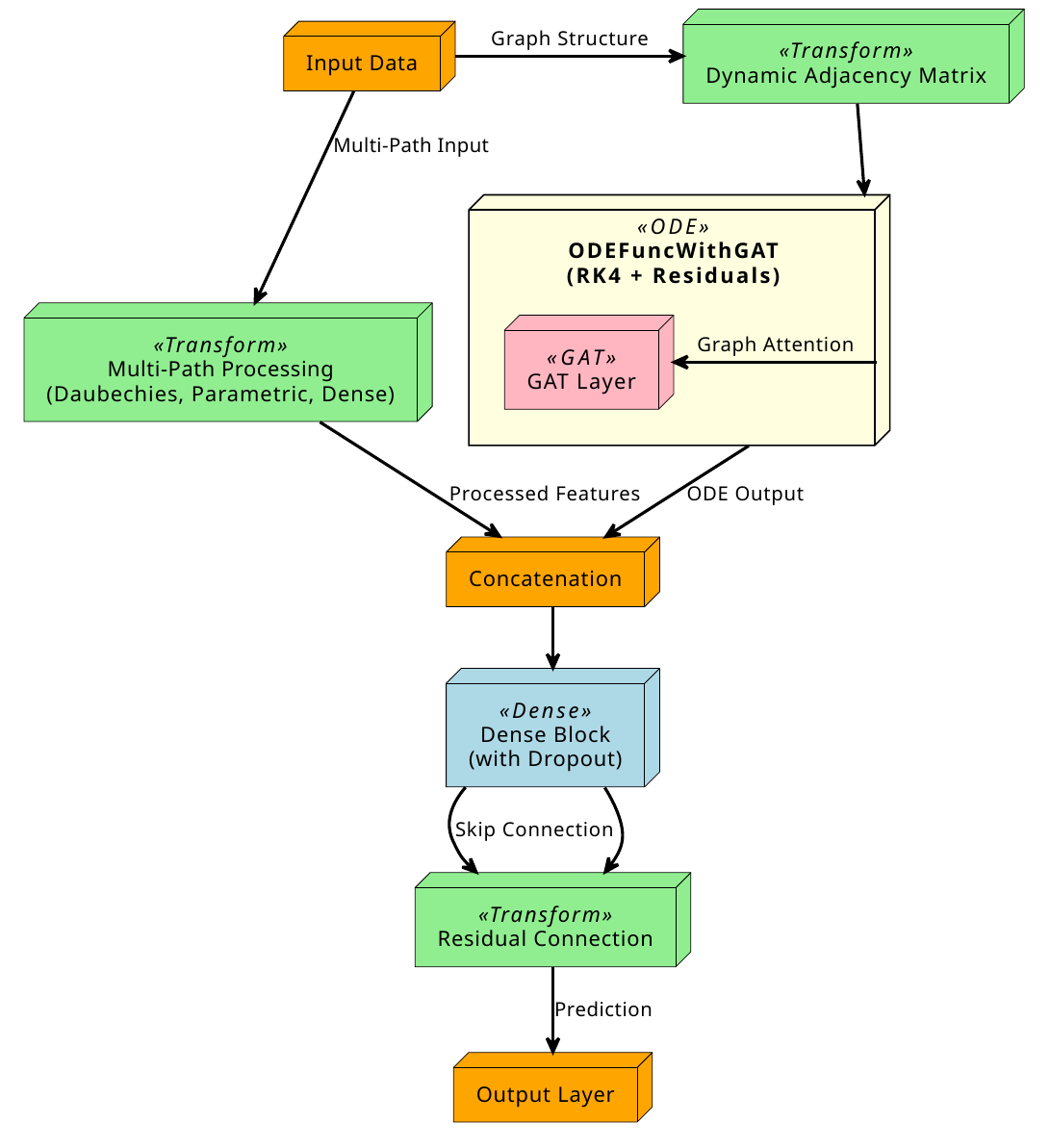}
	\caption{Overall architecture of the proposed model, depicting the dynamic graph construction, parallel processing paths, feature concatenation and final prediction layers with residual connections.}
	\label{fig:model1}
\end{figure}

The outputs from these paths—each tailored to a unique data characteristic—are concatenated into a unified feature vector, approximately of size \( 2h \) (i.e., \( h + h/2 + h/2 + h/4 \)), reflecting contributions from all paths \cite{pm1, pm2}. This vector is then refined through a sequence of dense layers (with output sizes \( h \), \( h/2 \), and back to \( h \)), incorporating dropout to mitigate overfitting and a residual connection from the first dense layer to enhance gradient flow and training stability. The final output is a scalar prediction \( y \in \mathbb{R} \), suitable for regression tasks. This multi-path architecture, visualized in Figure~\ref{fig:model1}, ensures a holistic representation by integrating diverse feature types, making it robust against the multifaceted nature of time series data.

\subsection{Dynamic Graph Construction}

\begin{figure}[ht]
	\centering
	\includegraphics[width = 1\columnwidth ,height=7cm]{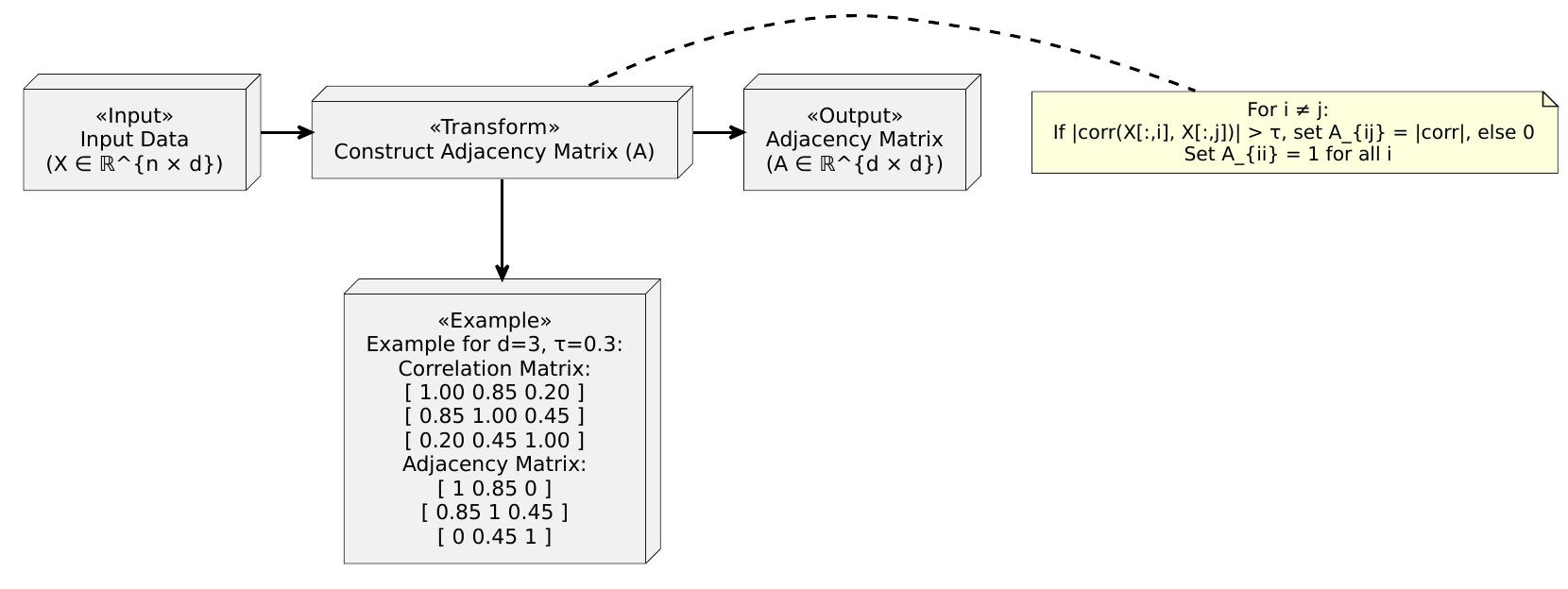}
	\caption{Dynamic adjacency matrix construction, showing feature connections based on correlation threshold \( \tau \).}
	\label{fig:adjacency}
\end{figure}

The dynamic graph construction is a pivotal component that generates an adjacency matrix based on feature correlations, enabling the graph attention mechanism to prioritize significant relationships. For an input dataset \( X \in \mathbb{R}^{n \times d} \), where \( n \) denotes the number of samples and \( d \) the number of features, the process begins by computing Pearson correlation coefficients for all feature pairs. These coefficients measure the linear relationship between features, providing a quantitative basis for connectivity. The adjacency matrix \( A \in \mathbb{R}^{d \times d} \) is then constructed as:
\begin{equation}
	A_{ij} = 
	\begin{cases} 
		|\text{corr}(X[:,i], X[:,j])| & \text{if } |\text{corr}(X[:,i], X[:,j])| > \tau \text{ and } i \neq j, \\
		1 & \text{if } i = j, \\
		0 & \text{otherwise},
	\end{cases}
\end{equation}
where \( \tau \) (e.g., 0.3) acts as a threshold to filter out weak correlations, ensuring only strong relationships form edges, and self-loops (\( A_{ii} = 1 \)) allow each feature to retain its own information in the attention mechanism. This construction, illustrated in Figure~\ref{fig:adjacency}, adapts dynamically to the dataset, unlike static graphs that rely on predefined topologies \cite{pm6}.

The rationale for this approach lies in its ability to reflect the inherent structure of the data, which is often unknown a priori in time series applications. For instance, in financial or biological datasets, feature dependencies may shift over time or vary across contexts; a static graph would fail to capture such nuances, whereas this method adjusts the graph topology based on empirical evidence. This adaptability enhances the model’s capacity to model complex, data-specific interactions, a critical factor in achieving accurate predictions.

The choice of Pearson correlation for constructing the adjacency matrix is motivated by its simplicity and computational efficiency, as it effectively captures linear relationships between features, which are prevalent in many time series datasets, including those addressed by similarity graph-correlation networks \cite{editor_paper_1} and heterogeneous graph convolution methods \cite{editor_paper_2}. However, Pearson correlation is limited to linear dependencies and may not capture nonlinear relationships. Alternative measures, such as mutual information or distance correlation, could be explored in future work to capture more complex dependencies. Additionally, in time-varying contexts where feature relationships change over time, the static correlation computed over the entire dataset may not fully capture these dynamics. To address this, one could consider computing correlations over sliding windows or incorporating time-dependent graph structures, although this would increase computational complexity.

With the adjacency matrix \( A \) defined, it guides the Neural ODE component, which leverages graph attention to model feature dynamics, as described next.

\subsection{Neural ODE Component with Graph Attention}

The Neural ODE component models the continuous-time evolution of features and their interactions, defined by the differential equation:
\begin{equation}
	\frac{dh}{dt} = f_\theta(h, A),
\end{equation}
where \( h \in \mathbb{R}^{d \times h} \) is the state, \( A \in \mathbb{R}^{d \times d} \) is the adjacency matrix, and \( f_\theta \) is a neural network comprising two dense layers (for feature transformation), a graph attention layer (to weigh feature interactions using \( A \)), residual connections (to stabilize training), and layer normalization (to normalize outputs). The graph attention mechanism dynamically adjusts the influence of each feature based on its correlations, enhancing the model’s relational awareness \cite{pm7}.

The graph attention mechanism within \( f_\theta \) operates by assigning attention weights to the features based on their correlations, as encoded in the adjacency matrix \( A \). Specifically, for each feature node, the attention mechanism computes a weighted sum of its neighbors' representations, where the weights are learned based on the similarity between the nodes' current states. This allows the model to dynamically focus on the most relevant feature interactions at each step of the state evolution. By integrating this into the Neural ODE, the model can capture both the temporal dynamics and the relational dependencies among features in a continuous-time framework.

The ODE is solved using the fourth-order Runge-Kutta (RK4) method \cite{pm9} with a step size \( dt = 0.01 \). The fourth-order Runge-Kutta (RK4) method is chosen for its high accuracy and efficiency, providing a fourth-order approximation with four evaluations per step, which balances precision and computational cost. The RK4 update is given by:
\begin{align}
	k_1 &= f_\theta(h), \\
	k_2 &= f_\theta(h + 0.5 \cdot dt \cdot k_1), \\
	k_3 &= f_\theta(h + 0.5 \cdot dt \cdot k_2), \\
	k_4 &= f_\theta(h + dt \cdot k_3),
\end{align}
and the state is updated as:
\begin{equation}
	h_t = h + \frac{dt}{6} (k_1 + 2k_2 + 2k_3 + k_4).
\end{equation}

Given that only a single integration step is performed, the choice of RK4 provides a high-accuracy approximation of the state evolution, with a global error of \( O(dt^4) \), which is sufficient for capturing the dynamics within this short horizon. The step size \( dt = 0.01 \) was selected based on preliminary experiments to balance computational efficiency and accuracy, though a more systematic analysis of sensitivity to \( dt \) could be conducted. Since the Neural ODE framework inherently supports continuous-time modeling, it is well-suited for handling irregularly sampled data by evaluating the state at arbitrary time points, thereby avoiding issues of convergence or drift associated with discrete-time models.

\begin{figure}[ht]
	\centering
	\includegraphics[width=1\columnwidth,height=8cm]{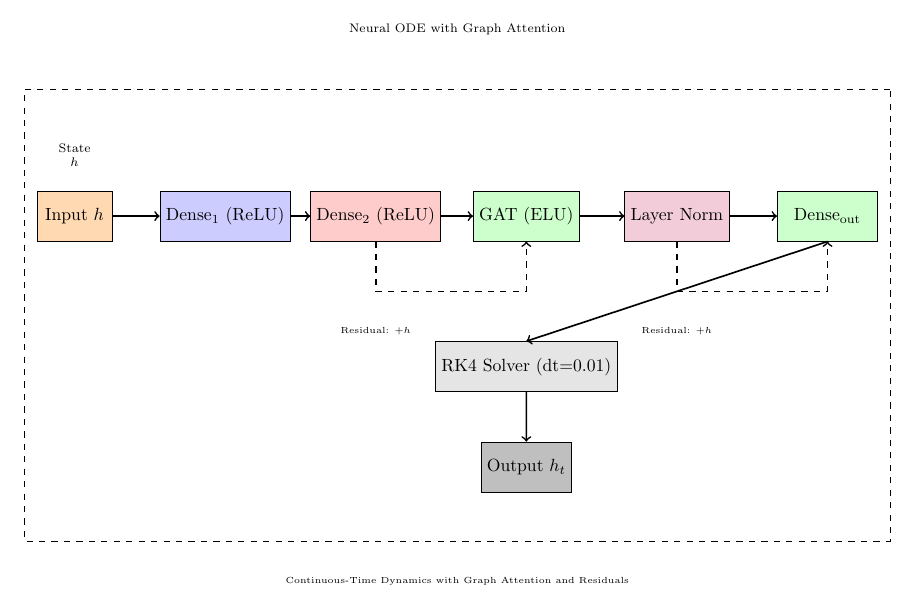}
	\caption{Visualization of the Neural ODE component with graph attention, showing \( f_\theta \) and the RK4 solver.}
	\label{fig:neural_ode}
\end{figure}

This process, shown in Figure~\ref{fig:neural_ode}, balances precision with computational efficiency. The output is pooled to \( h_{\text{ode}} \in \mathbb{R}^{h} \).

The Neural ODE is chosen for its ability to model dynamics continuously, avoiding the fixed-step limitations of discrete models like RNNs, which struggle with irregular sampling. The integration of graph attention further justifies its inclusion, as it leverages the dynamic graph to enhance feature interaction modeling, critical for interdependent time series.

While other graph neural network architectures, such as graph convolutional networks or graph transformers, could be considered, the graph attention mechanism is particularly well-suited for dynamic graph construction based on feature correlations. This mechanism enables adaptive weighting of feature interactions, which is essential given the data-driven nature of the adjacency matrix. Integrating transformers could potentially enhance the ability to capture long-range dependencies but it may introduce additional complexity without clear benefits for the current problem setup.

To complement the continuous dynamics of the Neural ODE, additional transform layers extract static temporal features, detailed below.

\subsection{Transform Layers}

To enrich the feature representation, the model includes two specialized transform layers—parametric and Daubechies-inspired—each designed to address distinct temporal properties, complementing the relational insights from the dynamic graph.

\subsubsection{Parametric Transform}

\begin{figure}[ht]
	\centering
	\includegraphics[width=1\columnwidth, height=8cm]{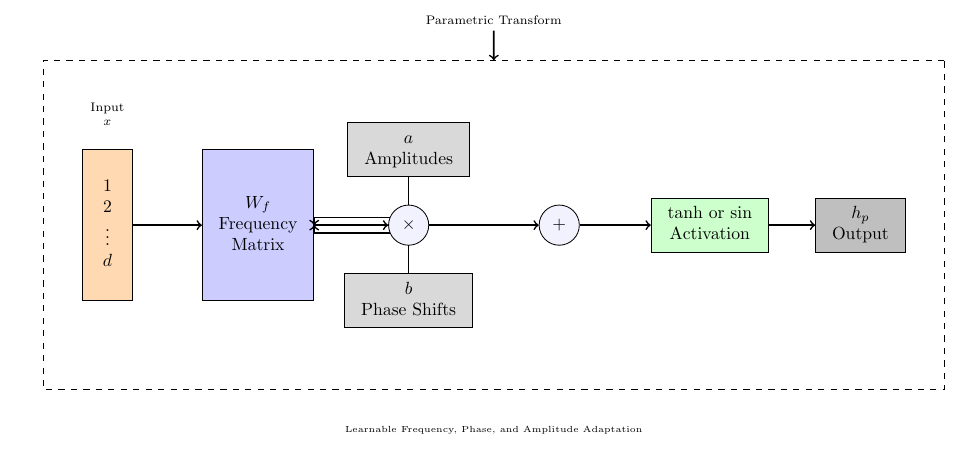}
	\caption{Process flow of the parametric transform, showing the learnable frequency matrix \( W_f \), phase shifts \( b \), amplitudes \( a \), and activation function.}
	\label{fig:model2}
\end{figure}

The parametric transform is engineered to capture periodic patterns within the time series through a flexible, learnable mechanism. For an input \( x \in \mathbb{R}^{d} \), it computes:
\begin{equation}
	h_p = 
	\begin{cases}
		\tanh(W_f x \cdot a + b) & \text{if activation is hyperbolic tangent}, \\
		\sin(W_f x \cdot a + b) & \text{otherwise},
	\end{cases}
\end{equation}
where \( W_f \in \mathbb{R}^{d \times h/2} \) is a trainable frequency matrix that maps the input to a latent space, \( a \in \mathbb{R}^{h/2} \) modulates the amplitude of the resulting signals, and \( b \in \mathbb{R}^{h/2} \) adjusts phase shifts, with \( h \) as the hidden dimension and the output size halved. The operation involves a matrix multiplication followed by element-wise scaling and shifting, with the activation function (\(\tanh\) or \(\sin\)) determining the shape of the periodic response. This process, depicted in Figure~\ref{fig:model2}, allows the transform to adapt to the data’s specific periodicities during training.

This component is chosen over traditional methods like Fourier transforms because it learns frequency patterns directly from the data, rather than assuming fixed basis functions. For example, in datasets with irregular or evolving cycles (e.g., seasonal trends in climate data), a fixed-frequency approach might miss subtle variations, whereas this transform’s adaptability ensures precise modeling of such behaviors, enhancing predictive accuracy \cite{pm3,pm4}.

\subsubsection{Daubechies Transform}

The Daubechies transform focuses on extracting multi-scale temporal features, drawing inspiration from Daubechies wavelets, which are renowned for their ability to analyze localized signal variations \cite{pm5}. It operates on \( x \in \mathbb{R}^{d} \) as:
\begin{equation}
	h_d = \text{GlobalAveragePool}\left( \text{ReLU}\left( \text{Conv1D}(x, K, \text{padding=SAME}) + b \right) \right),
\end{equation}
where \( K \in \mathbb{R}^{5 \times 1 \times h/2} \) is a convolutional kernel with a size of 5 and \( h/2 \) filters, \( b \in \mathbb{R}^{h/2} \) is a bias term, and "SAME" padding ensures the output matches the input’s dimensionality before pooling. The Daubechies transform applies a 1D convolution with a learnable kernel to capture multi-scale temporal features, followed by ReLU activation and global average pooling.

\begin{figure}[ht]
	\centering
	\includegraphics[width=1\columnwidth, height=8cm]{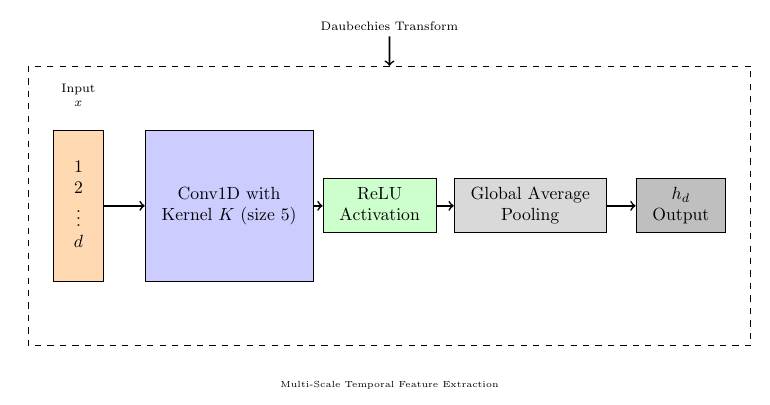}
	\caption{Illustration of the Daubechies transform, depicting convolution with kernel \( K \), ReLU activation, and global average pooling.}
	\label{fig:model3}
\end{figure}

The use of a wavelet-inspired transform is justified by its ability to capture multi-scale features and transients, which are crucial in time series with non-stationary behavior or sudden changes. Unlike Fourier transforms, which assume stationary signals with fixed frequencies, wavelets provide localization in both time and frequency, making them more suitable for diverse time series applications. Figure~\ref{fig:model3} illustrates this workflow.

\subsection{Feature Integration and Forward Pass}

Having outlined the individual feature extraction paths, this subsection describes their integration into a cohesive prediction pipeline. The forward pass synthesizes features from all paths into a cohesive prediction. The Neural ODE path yields \( h_{\text{ode}} \in \mathbb{R}^{h} \), capturing temporal and relational dynamics; the Daubechies path produces \( h_d \in \mathbb{R}^{h/2} \), emphasizing multi-scale features; the parametric path generates \( h_p \in \mathbb{R}^{h/2} \), focusing on periodicities; and the dense path outputs \( h_{\text{dense}} \in \mathbb{R}^{h/4} \), retaining raw data information. These are concatenated:
\begin{equation}
	h_c = [h_{\text{ode}} \parallel h_d \parallel h_p \parallel h_{\text{dense}}],
\end{equation}
forming \( h_c \in \mathbb{R}^{2h} \) (approximately \( h + h/2 + h/2 + h/4 \)). This combined vector is processed through dense layers with sizes \( h \), \( h/2 \), and back to \( h \), incorporating dropout \cite{pm8} (e.g., rate 0.2) to prevent overfitting and a residual connection from the first dense layer to enhance training by preserving earlier features. The final dense layer outputs a scalar \( y \in \mathbb{R} \).

This integration is essential for unifying diverse feature representations, ensuring no single aspect (e.g., periodicity or transients) dominates the prediction \cite{pm10}. The residual connection addresses the vanishing gradient problem, a common issue in deep networks, making this design both robust and trainable.

With the forward pass defined, the model is trained using an optimized methodology, as follows.

\subsection{Training Methodology}

Training minimizes the mean squared error using the Adam optimizer, which adapts learning rates for efficient convergence. Hyperparameters (e.g., learning rate, hidden dimension) are tuned via Bayesian optimization to balance performance and complexity \cite{pm11}. Early stopping halts training when validation loss plateaus, and learning rate reduction adjusts the step size dynamically, ensuring stability across varied datasets. This methodology is selected for its proven efficacy in optimizing deep, complex models while preventing overfitting.

The training process leverages the synergistic roles of these components, elaborated next.

\subsection{Information Fusion Across Components}

The proposed model fuses heterogeneous features from multiple paths, aligning with multi-source information fusion principles. The dynamic graph construction extracts relational information via an adjacency matrix \( A \in \mathbb{R}^{d \times d} \) from feature correlations in \( X \in \mathbb{R}^{n \times d} \), guiding subsequent fusion. The Neural ODE with graph attention fuses this with continuous-time dynamics, yielding \( h_{\text{ode}} \in \mathbb{R}^{h} \) that integrates feature interdependencies and temporal evolution. The Daubechies transform fuses multi-scale transients into \( h_d \in \mathbb{R}^{h/2} \), while the parametric transform captures adaptive periodicities in \( h_p \in \mathbb{R}^{h/2} \). The dense path preserves raw data as \( h_{\text{dense}} \in \mathbb{R}^{h/4} \). These outputs are concatenated into \( h_c \in \mathbb{R}^{2h} \) and refined with residual connections, achieving decision-level fusion for a scalar prediction \( y \in \mathbb{R} \). This synergy, shown in Figures~\ref{fig:model1}–\ref{fig:model3}, enhances robustness for time series applications like sensor or financial data analysis. This integrated design incurs a computational cost, analyzed below.

\subsection{Computational Complexity}

Complexity is evaluated with \( d \) as features, \( n \) as samples, and \( h \) as hidden dimension. Dynamic graph construction requires \( O(d^2 \cdot n) \) for correlations and \( O(d^2) \) for thresholding. Parametric and Daubechies transforms are \( O(d \cdot h) \) due to matrix operations and convolutions. The Neural ODE with graph attention scales as \( O(d^2 \cdot h) \) per RK4 step, driven by the graph attention mechanism applied to the adjacency matrix \( A \). Final dense layers are \( O(h^2) \). Total complexity is \( O(d^2 \cdot n + d^2 \cdot h) \), with space complexity \( O(d^2 + d \cdot h) \) from \( A \) and intermediate features, practical with optimized implementations.

\subsection{Algorithm}

The prediction process of the proposed model is formalized below, integrating dynamic graph construction, multi-path feature extraction, and final processing into a clear and concise workflow. This algorithm outlines the essential steps to transform input data into a scalar prediction, leveraging each component’s strengths.

\begin{algorithm}[!htbp]
	\caption{Prediction Process of the Proposed Model}
	\label{alg:proposed_model}
	\begin{algorithmic}[1]
		\Require Input data \( X \in \mathbb{R}^{n \times d} \), input vector \( x \in \mathbb{R}^{d} \), threshold \( \tau \), hidden dimension \( h \)
		\Ensure Prediction \( y \in \mathbb{R} \)
		\State Calculate feature correlations for \( X \) and build adjacency matrix \( A \in \mathbb{R}^{d \times d} \), setting \( A_{ij} = |\text{corr}(X[:,i], X[:,j])| \) if above \( \tau \) and \( i \neq j \), \( A_{ii} = 1 \), otherwise \( A_{ij} = 0 \)
		\State Project \( x \) to a higher-dimensional state \( h_0 \in \mathbb{R}^{d \times h} \) using a dense layer
		\State Evolve \( h_0 \) through a Neural ODE with graph attention based on \( A \), using a single RK4 step (\( dt = 0.01 \)), and pool to obtain \( h_{\text{ode}} \in \mathbb{R}^{h} \)
		\State Apply Daubechies transform to \( x \), using convolution and pooling, to get \( h_d \in \mathbb{R}^{h/2} \)
		\State Compute parametric transform on \( x \), adapting frequencies, to produce \( h_p \in \mathbb{R}^{h/2} \)
		\State Process \( x \) through a dense layer to yield \( h_{\text{dense}} \in \mathbb{R}^{h/4} \)
		\State Concatenate features into \( h_c = [h_{\text{ode}} \parallel h_d \parallel h_p \parallel h_{\text{dense}}] \in \mathbb{R}^{2h} \)
		\State Refine \( h_c \) through dense layers with dropout and a residual connection
		\State Generate scalar prediction \( y \) via a final dense layer
		\State \Return \( y \)
	\end{algorithmic}
\end{algorithm}

This simplified algorithm encapsulates the model’s core operations, ensuring accessibility while preserving the integrity of the prediction process.

\section{Data Description and Preprocessing}

This study employs seven time series datasets to evaluate the proposed approach: the ETTh1, ETTh2, ETTm1, and ETTm2 datasets \cite{ett_dataset}, which provide electricity transformer temperature readings, and the monthly Waste Energy Consumption, Hydroelectric Power Consumption, and Solar Energy Consumption datasets from the U.S. Energy Information Administration (EIA) \cite{eia_dataset}. These datasets were selected for their diverse temporal dynamics, including high-frequency (hourly and 15-minute intervals) and low-frequency (monthly) patterns, making them suitable benchmarks for assessing multi-resolution forecasting capabilities.

The ETTh1 and ETTh2 datasets each contain 17,420 hourly observations spanning from July 1, 2016, 00:00, to June 26, 2018, 19:00, while the ETTm1 and ETTm2 datasets each include 69,680 observations at 15-minute intervals over the same period, ending at 19:45 on June 26, 2018. These datasets capture the target variable (Oil Temperature, OT) along with six power load metrics, offering a multivariate structure suitable for analyzing short-term and long-term dependencies in energy monitoring applications. Anomalies are addressed using the Interquartile Range (IQR) method \cite{df1} to ensure target values remain within realistic ranges, and incomplete or corrupted timestamp records are removed to maintain data integrity.

The EIA datasets—Waste, Hydro, and Solar—provide monthly energy consumption data. The Waste and Hydro datasets each contain 617 observations from January 1973 to May 2024, while the Solar dataset includes 485 observations from January 1984 to May 2024. These datasets originally include columns such as MSN (object), YYYYMM (int64), Value (object), Column\_Order (int64), Description (object), and Unit (object) with Value representing energy consumption in trillion Btu. After preprocessing, only the target variable (energy consumption as float64) is retained before resampling to monthly frequency. These datasets are critical for studying seasonal trends and long-term growth patterns in energy consumption, with no explicit outlier handling applied due to their aggregated nature.

Preprocessing involves converting date columns to a standardized datetime format and setting them as the index. For the EIA datasets, missing or non-numeric entries in the raw data are dropped, and missing values from lagged or rolling features are imputed using forward-fill and backward-fill methods \cite{forwardbackward}. For the ETT datasets, similar imputation techniques address gaps from lagging and rolling operations. After preprocessing, the EIA datasets each have 14 features (e.g., month, quarter, lags, rolling statistics, Fourier terms), while the ETT datasets have 51 features (including extensive time-based, lag, rolling, and expanding features). All features are normalized using a Standard Scaler \cite{standardscaler} to standardize their scale, ensuring equitable contribution to the forecasting process.

Table \ref{tab:dataset_info} summarizes the key characteristics of each dataset after preprocessing.

\begin{table}[h]
	\centering
	\caption{Datasets Information}
	\label{tab:dataset_info}
	\begin{tabular}{|l|c|c|c|c|c|}
		\hline
		\text{Dataset} & \text{Observations} & \text{Start Date} & \text{End Date} & \text{Frequency} & \text{Features} \\
		\hline
		\text{Waste} & 617 & 1973-01 & 2024-05 & \text{monthly} & 14 \\
		\text{Hydro} & 617 & 1973-01 & 2024-05 & \text{monthly} & 14 \\
		\text{Solar} & 485 & 1984-01 & 2024-05 & \text{monthly} & 14 \\
		\text{ETTh1} & 17420 & 2016-07-01 00:00 & 2018-06-26 19:00 & \text{hourly} & 51 \\
		\text{ETTh2} & 17420 & 2016-07-01 00:00 & 2018-06-26 19:00 & \text{hourly} & 51 \\
		\text{ETTm1} & 69680 & 2016-07-01 00:00 & 2018-06-26 19:45 & \text{15min} & 51 \\
		\text{ETTm2} & 69680 & 2016-07-01 00:00 & 2018-06-26 19:45 & \text{15min} & 51 \\
		\hline
	\end{tabular}
\end{table}

\section{Feature Representation and Extraction}

To capture temporal dependencies, seasonality, and statistical properties \cite{df2}, feature engineering is applied across all datasets. For the monthly EIA datasets (Waste, Hydro, Solar), temporal features including \textit{month}, \textit{quarter}, and \textit{day\_of\_year} are extracted as:
\begin{equation}
	\mathbf{T}_{\text{EIA}} = \{\text{month}, \text{quarter}, \text{day\_of\_year}\},
\end{equation}
representing periodic structures within the data. Lagged dependencies are captured through autoregressive components \cite{df3} with lags from the previous six time steps:
\begin{equation}
	x_{t,k} = y_{t-k}, \quad k = 1, \ldots, 6,
\end{equation}
Rolling statistics, including mean, standard deviation, and median, are computed using a window size of three \cite{df4}:
\begin{equation}
	\begin{aligned}
		\mu_{t,3} &= \frac{1}{3}\sum_{i=t-2}^{t} y_i, \\
		\sigma_{t,3} &= \sqrt{\frac{1}{3} \sum_{i=t-2}^{t} (y_i - \mu_{t,3})^2}, \\
		\text{med}_{t,3} &= \text{median}(y_{t-2}, \ldots, y_t).
	\end{aligned}
\end{equation}
Seasonality \cite{df5} is captured through Fourier components \cite{df6}, representing monthly cyclical patterns as:
\begin{equation}
	\begin{aligned}
		F_{\sin}(t) &= \sin\left(\frac{2\pi t}{12}\right), \\
		F_{\cos}(t) &= \cos\left(\frac{2\pi t}{12}\right),
	\end{aligned}
\end{equation}
where $t$ is the month index.

\begin{figure}[h]
	\centering
	\includegraphics[width=1\textwidth]{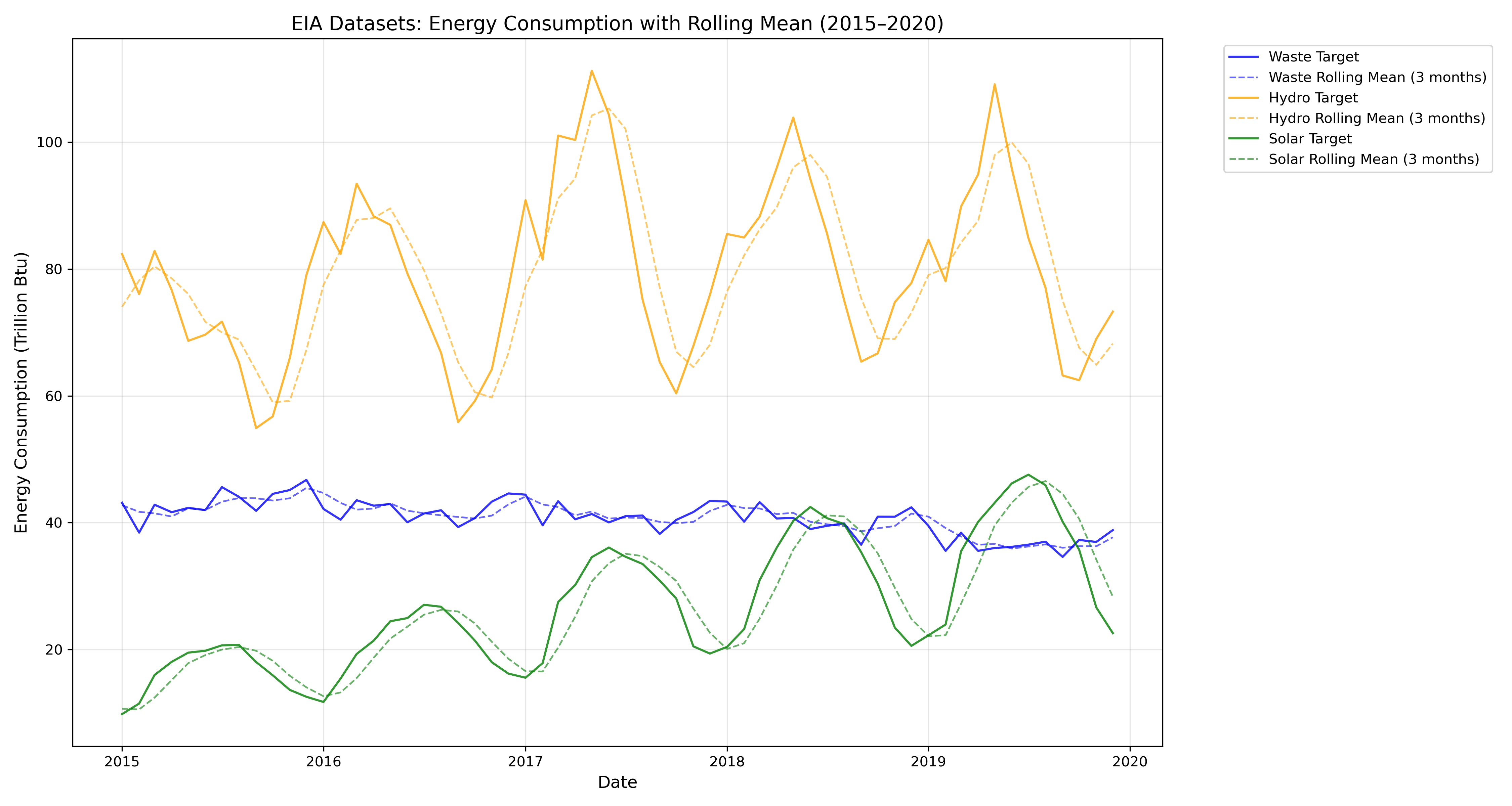}
	\caption{EIA Datasets: Energy Consumption with Rolling Mean (2015-2020) for Waste, Hydro, and Solar.}
	\label{fig:eia_energy_consumption}
\end{figure}

For the hourly ETT datasets (ETTh1, ETTh2, ETTm1, ETTm2), a comprehensive set of features is engineered. Temporal features include \textit{hour}, \textit{day}, \textit{day\_of\_week}, \textit{month}, \textit{quarter}, \textit{day\_of\_year}, \textit{week\_of\_year}, and an indicator for weekends (\textit{is\_weekend}), represented as:
\begin{equation}
	\mathbf{T}_{\text{ETT}} = \{\text{hour}, \text{day}, \text{day\_of\_week}, \text{month}, \text{quarter}, \text{day\_of\_year}, \text{week\_of\_year}, \text{is\_weekend}\}.
\end{equation}
Cyclical encoding is applied to hour, day, and month using sine and cosine transformations:
\begin{equation}
	\begin{aligned}
		F_{\sin}^{\text{hour}}(t) &= \sin\left(\frac{2\pi t}{24}\right), & F_{\cos}^{\text{hour}}(t) &= \cos\left(\frac{2\pi t}{24}\right), \\
		F_{\sin}^{\text{day}}(t) &= \sin\left(\frac{2\pi t}{31}\right), & F_{\cos}^{\text{day}}(t) &= \cos\left(\frac{2\pi t}{31}\right), \\
		F_{\sin}^{\text{month}}(t) &= \sin\left(\frac{2\pi t}{12}\right), & F_{\cos}^{\text{month}}(t) &= \cos\left(\frac{2\pi t}{12}\right).
	\end{aligned}
\end{equation}
Lagged features are included for specific lags \cite{df3}:
\begin{equation}
	x_{t,k} = y_{t-k}, \quad k \in \{1, 2, 3, 5, 7, 14, 21, 28\},
\end{equation}
Rolling statistics, including mean, standard deviation, minimum, maximum, variance, and skewness, are computed for multiple window sizes \cite{df4}:
\begin{equation}
	w \in \{5, 10, 20\},
\end{equation}
with statistical features such as:
\begin{equation}
	\begin{aligned}
		\mu_{t,w} &= \frac{1}{w}\sum_{i=t-w+1}^{t} y_i, \\
		\sigma_{t,w} &= \sqrt{\frac{1}{w} \sum_{i=t-w+1}^{t} (y_i - \mu_{t,w})^2}, \\
		\text{min}_{t,w} &= \min(y_{t-w+1}, \ldots, y_t), \\
		\text{max}_{t,w} &= \max(y_{t-w+1}, \ldots, y_t), \\
		\text{var}_{t,w} &= \frac{1}{w} \sum_{i=t-w+1}^{t} (y_i - \mu_{t,w})^2, \\
		\text{skew}_{t,w} &= \frac{\frac{1}{w} \sum_{i=t-w+1}^{t} (y_i - \mu_{t,w})^3}{\left(\frac{1}{w} \sum_{i=t-w+1}^{t} (y_i - \mu_{t,w})^2\right)^{3/2}}.
	\end{aligned}
\end{equation}
Additionally, expanding window statistics (mean, standard deviation, minimum, and maximum) are computed to capture long-term trends:
\begin{equation}
	\begin{aligned}
		\mu_{t,\text{exp}} &= \frac{1}{t}\sum_{i=1}^{t} y_i, \\
		\sigma_{t,\text{exp}} &= \sqrt{\frac{1}{t} \sum_{i=1}^{t} (y_i - \mu_{t,\text{exp}})^2}, \\
		\text{min}_{t,\text{exp}} &= \min(y_1, \ldots, y_t), \\
		\text{max}_{t,\text{exp}} &= \max(y_1, \ldots, y_t).
	\end{aligned}
\end{equation}

\begin{figure}[h]
	\centering
	\includegraphics[width=1\textwidth,height=8cm]{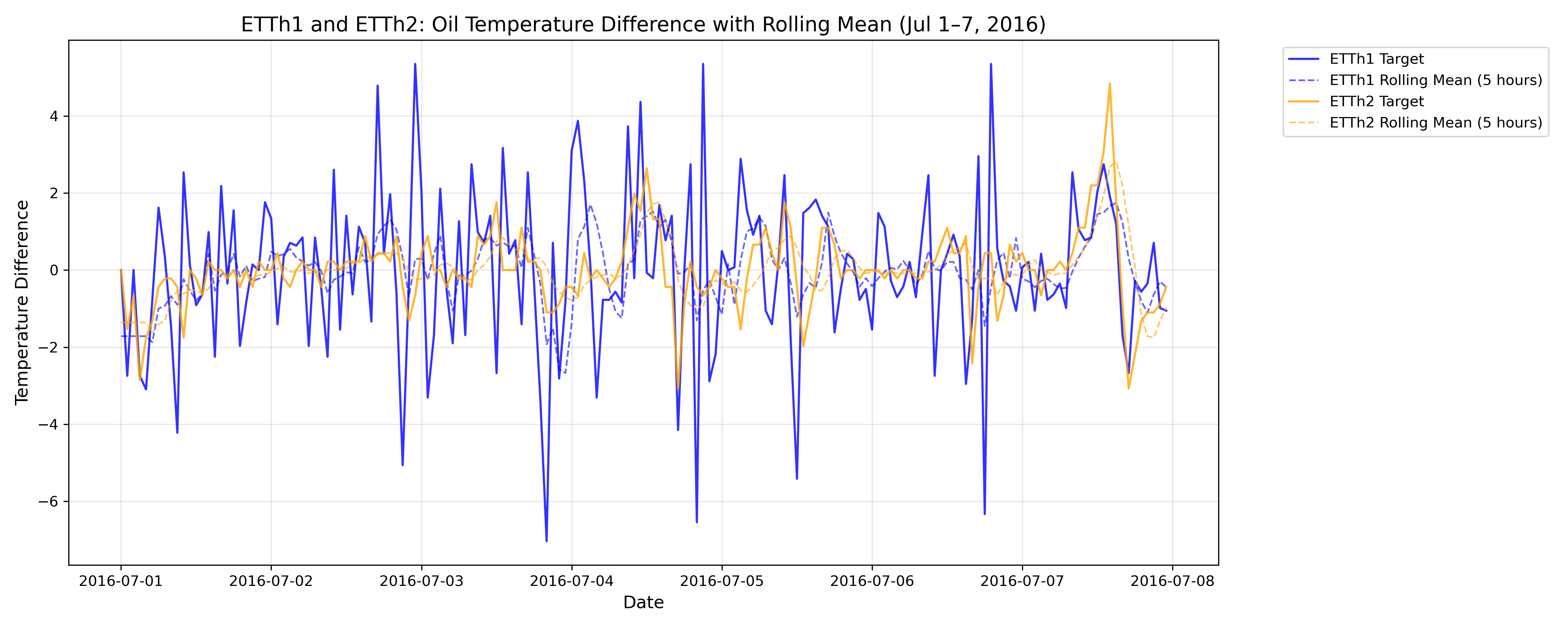}
	\caption{ETTh1 and ETTh2: Oil Temperature Difference with Rolling Mean (Jul 1-7, 2016).}
	\label{fig:etth1_etth2_temperature_difference}
\end{figure}

\begin{figure}[h]
	\centering
	\includegraphics[width=1\textwidth, height=8cm]{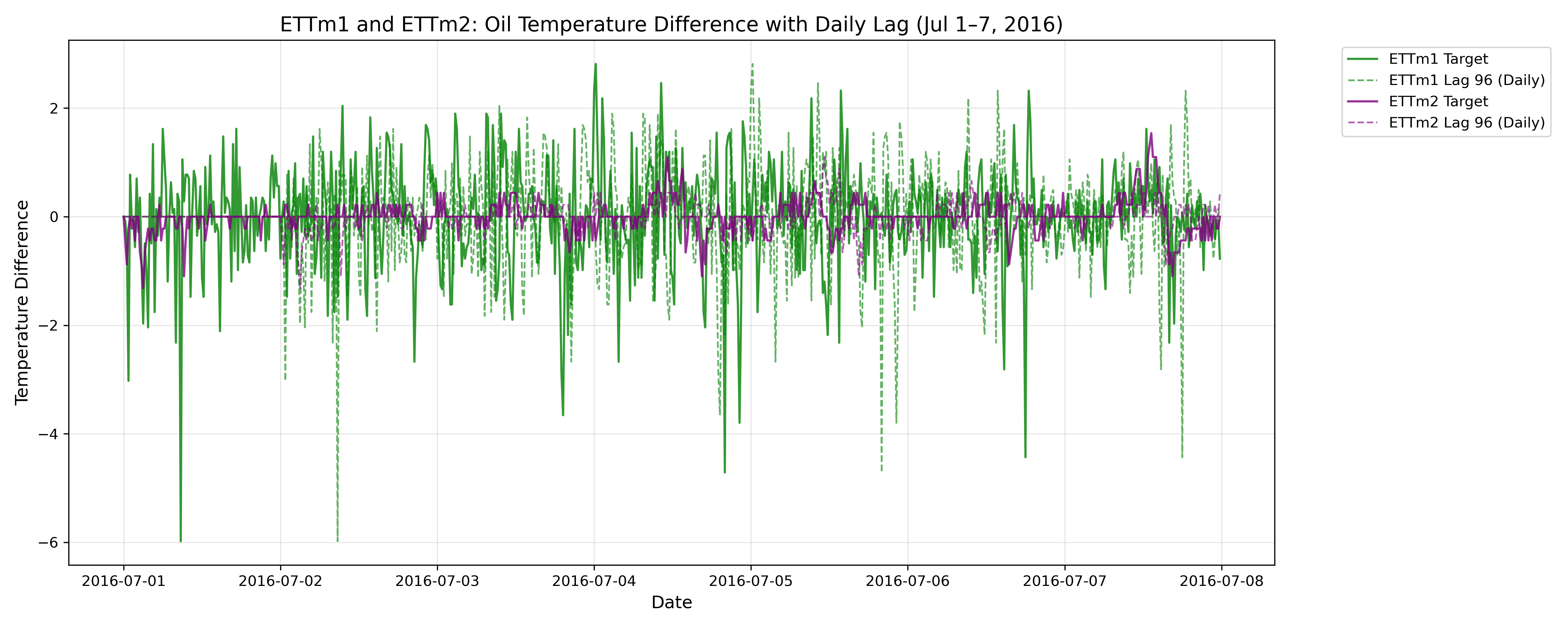}
	\caption{ETTm1 and ETTm2: Oil Temperature Difference with Daily Lag (Jul 1-7, 2016).}
	\label{fig:ettm1_ettm2_temperature_difference}
\end{figure}

For the ETT datasets (ETTh1, ETTh2, ETTm1, ETTm2), the forecasting framework accommodates two distinct input structures: (1) sequences of past target variable observations with varying lengths ([96, 120, 336, 720] time steps), enabling the capture of temporal dependencies at multiple scales and (2) the engineered features described above. For the EIA datasets (Waste, Hydro, Solar), only the engineered features are used with a feature selection process. 

\subsection{Feature Selection Methodology}

Feature selection follows a two-stage process to optimize feature relevance \cite{df7}. First, an initial ranking of features is performed based on their importance, followed by a refinement step using Recursive Feature Elimination (RFE) \cite{df9}. In the initial stage, feature importance is calculated as:
\begin{equation}
	I(f_i) = \sum_{j=1}^T \text{gain}_j(f_i),
\end{equation}
where $I(f_i)$ represents the cumulative importance of feature $i$ over $T$ iterations. The top 15 features are selected for further refinement. In the second stage, RFE iteratively removes less informative features until only eight remain, defining the optimal subset as:
\begin{equation}
	S^* = \arg\min_{S \subset S_0} \text{MSE}(S) \quad \text{s.t. } |S| = 8,
\end{equation}
where $S_0$ is the initial feature set from the first stage. for EIA datasets, which contain exactly 14 features, RFE is applied directly to select the top 8 features, bypassing the initial ranking step.

\section{Experimental Setup}

The experiments were conducted under two different setups depending on the evaluation type. All multi-window evaluations were performed on Kaggle using a P100/T4 GPU, ensuring sufficient computational resources for training complex time series models. In contrast, the non-windowed evaluations were conducted on a local workstation equipped with an Intel i7-12700 processor and 32 GB of RAM. All datasets used in this study are multivariate and were split into 80\% for training and 20\% for testing.

\subsection{Baseline Model Comparison}

The performance of the proposed model was evaluated through a comparative analysis against established baseline models. In the experiments with the ETT datasets, the proposed model was compared with baseline models including N-BEATS \cite{nbeats}, N-HiTS \cite{nhits}, and TCN \cite{tcn}. For the renewable energy datasets, the evaluation involved benchmarks such as XGBoost \cite{xgboost}, RandomForest \cite{rf}, Attention \cite{attention}, LSTM+Attention (LSTM-Attn), N-BEATS, TCN, RNN-GRU, and Informer \cite{informer}.

\subsection{Hyperparameter Tuning}

Hyperparameter optimization was performed using Optuna \cite{optuna} across all models and datasets. For the proposed Neural ODE with Dynamic GAT, the Waste and Hydro datasets tuned hidden dimensions (16--512), batch size (\{16, 32\}), epochs (30--250), dropout rate (0.1--0.5, step 0.1), and learning rate ($10^{-4}$--0.01, logarithmic). The Solar dataset excluded dropout, tuning hidden dimensions (16--256), batch size (\{16, 32, 64, 128\}), epochs (30--150), and learning rate ($10^{-4}$--0.01, logarithmic). For ETTh1, ETTh2, ETTm1, and ETTm2, ranges included hidden dimensions (32--128, step 16), batch size (\{16, 32, 64, 128\}), epochs (30--100, step 10), dropout rate (0.1--0.5, step 0.1), and learning rate ($10^{-4}$--$10^{-2}$, logarithmic). Early stopping (patience=15) and learning rate reduction (factor=0.2, patience=5) were applied.

Traditional models included XGBoost with number of estimators (50--1000), maximum depth (3--12), and learning rate (0.01--0.3, logarithmic), and Random Forest with number of estimators (50--500), maximum depth (3--25), and maximum features (\{`sqrt', `log2', None\}). Deep models (20 trials) tuned Simple Attention with units (32--256), dropout (0.1--0.5), and learning rate ($10^{-4}$--$10^{-2}$, logarithmic), and LSTM+Attention with LSTM units (32--256) and dense units (16--128). Darts models optimized N-BEATS with input chunk length (10--50) and layer widths (128--512), TCN with number of filters (16--128) and kernel size (2--5), and RNN-GRU with hidden dimensions (32--256) and number of RNN layers (1--3). Informer tuned encoder layers (2--6), decoder layers (1--3), attention heads (4--12), and learning rate ($10^{-4}$--$10^{-2}$, logarithmic) for renewable energy datasets. All optimizations minimized mean squared error (MSE) on validation splits, using early stopping and learning rate reduction where applicable.

\section{Results and Discussion}

This section presents the performance of the proposed model, evaluated on ETT and Renewable energy datasets, with results organized separately for non-windowed and windowed configurations. Model performance is assessed using standard regression metrics, including MSE, RMSE, MAE, and $R^2$, which measure prediction accuracy and reliability. Lower MSE, RMSE, and MAE indicate reduced errors, while higher $R^2$ signifies better explanatory power. For multi-window settings, only RMSE and MAE are reported, as these metrics directly quantify absolute and scale-dependent errors, making them more interpretable for evaluating performance across different forecasting horizons.

\subsection{Results and Discussion for ETT Datasets}

\begin{table}[htbp]
	\centering
	\caption{Performance Comparison on ETT Datasets in the Non-Window Setting. Models with a window size of ‘None’ operate on continuous time series data without fixed-size temporal windows.}
	\label{tab:ett_results_1}
	\begin{tabular}{lllccccc}
		\toprule
		Dataset & Setting        & Model      & Window & MSE    & RMSE   & MAE    & $R^2$  \\
		\midrule
		\multicolumn{8}{c}{\textbf{Non-window Setting}} \\
		\midrule
		ETTh1   & Non-window     & Proposed   & --   & \textbf{0.0135} & \textbf{0.1161} & \textbf{0.0843} & \textbf{0.9606} \\
		ETTh1   & Non-window     & N-BEATS    & --   & 0.4709 & 0.6862 & 0.5225 & -0.5858 \\
		ETTh1   & Non-window     & N-HiTS     & --   & 0.2970 & 0.5449 & 0.4168 & -0.0000 \\
		ETTh1   & Non-window     & TCN        & --   & 0.3450 & 0.5874 & 0.4574 & -0.1619 \\
		\midrule
		ETTh2   & Non-window     & Proposed   & --   & \textbf{0.0220} & \textbf{0.1483} & \textbf{0.1076} & \textbf{0.9825} \\
		ETTh2   & Non-window     & N-BEATS    & --   & 2.0011 & 1.4146 & 1.0704 & -0.4271 \\
		ETTh2   & Non-window     & N-HiTS     & --   & 1.4498 & 1.2041 & 0.9359 & -0.0339 \\
		ETTh2   & Non-window     & TCN        & --   & 1.3552 & 1.1641 & 0.9074 & 0.0335 \\
		\midrule
		ETTm1   & Non-window     & Proposed   & --   & \textbf{0.0007} & \textbf{0.0273} & \textbf{0.0199} & \textbf{0.9904} \\
		ETTm1   & Non-window     & N-BEATS    & --   & 0.0675 & 0.2599 & 0.2036 & -0.0362 \\
		ETTm1   & Non-window     & N-HiTS     & --   & 0.0653 & 0.2556 & 0.1978 & -0.0025 \\
		ETTm1   & Non-window     & TCN        & --   & 0.0652 & 0.2553 & 0.1983 & -0.0006 \\
		\midrule
		ETTm2   & Non-window     & Proposed   & --   & \textbf{0.0004} & \textbf{0.0192} & \textbf{0.0141} & \textbf{0.9941} \\
		ETTm2   & Non-window     & N-BEATS    & --   & 0.0627 & 0.2504 & 0.2048 & 0.0034 \\
		ETTm2   & Non-window     & N-HiTS     & --   & 0.0642 & 0.2534 & 0.1883 & -0.0211 \\
		ETTm2   & Non-window     & TCN        & --   & 0.0629 & 0.2507 & 0.2032 & 0.0005 \\
		\bottomrule
	\end{tabular}
\end{table}

\begin{longtable}{lcccccccc}
	\caption{Multivariate Long-term Prediction Results on ETT Datasets across various window sizes. The smaller the RMSE and MAE scores, the better the performance, with the best results highlighted in bold}.
	\label{tab:ett_results_2} \\
	\toprule
	Dataset & \multicolumn{2}{c}{Proposed} & \multicolumn{2}{c}{N-BEATS} & \multicolumn{2}{c}{N-HiTS} & \multicolumn{2}{c}{TCN} \\
	\cmidrule(lr){2-3} \cmidrule(lr){4-5} \cmidrule(lr){6-7} \cmidrule(lr){8-9}
	& RMSE & MAE & RMSE & MAE & RMSE & MAE & RMSE & MAE \\
	\midrule
	\endfirsthead
	\multicolumn{9}{c}{\tablename\ \thetable\ -- Continued from Previous Page} \\
	\toprule
	Dataset & \multicolumn{2}{c}{Proposed} & \multicolumn{2}{c}{N-BEATS} & \multicolumn{2}{c}{N-HiTS} & \multicolumn{2}{c}{TCN} \\
	\cmidrule(lr){2-3} \cmidrule(lr){4-5} \cmidrule(lr){6-7} \cmidrule(lr){8-9}
	& RMSE & MAE & RMSE & MAE & RMSE & MAE & RMSE & MAE \\
	\midrule
	\endhead
	\midrule
	\multicolumn{9}{r}{Continued on Next Page} \\
	\endfoot
	\bottomrule
	\endlastfoot
	
	\small
	ETTh1 (Window 96)  & \textbf{0.0731} & \textbf{0.0496} & 0.6334 & 0.4962 & 0.5474 & 0.4176 & 0.5449 & 0.4176 \\
	ETTh1 (Window 120) & \textbf{0.0640} & \textbf{0.0481} & 0.6097 & 0.4683 & 0.5534 & 0.4229 & 0.5553 & 0.4292 \\
	ETTh1 (Window 336) & \textbf{0.0586} & \textbf{0.0376} & 0.6121 & 0.4844 & 0.5473 & 0.4177 & 0.5555 & 0.4267 \\
	ETTh1 (Window 720) & \textbf{0.0695} & \textbf{0.0513} & 0.5472 & 0.4202 & 0.5475 & 0.4195 & 0.6045 & 0.4761 \\
	\midrule
	ETTh2 (Window 96)  & \textbf{0.2054} & \textbf{0.1388} & 1.1821 & 0.8957 & 0.8860 & 0.6816 & 1.1702 & 0.9057 \\
	ETTh2 (Window 120) & \textbf{0.1410} & \textbf{0.0912} & 0.8090 & 0.6099 & 0.9777 & 0.7417 & 1.1668 & 0.9046 \\
	ETTh2 (Window 336) & \textbf{0.1086} & \textbf{0.0596} & 1.3264 & 1.0088 & 1.3202 & 1.0329 & 1.1683 & 0.8984 \\
	ETTh2 (Window 720) & \textbf{0.1453} & \textbf{0.0941} & 0.9265 & 0.6780 & 1.1589 & 0.8619 & 1.2791 & 0.9931 \\
	\midrule
	ETTm1 (Window 96)  & \textbf{0.0375} & \textbf{0.0264} & 0.2584 & 0.2023 & 0.2556 & 0.1978 & 0.2553 & 0.1983 \\
	ETTm1 (Window 120) & \textbf{0.0533} & \textbf{0.0441} & 0.2610 & 0.2046 & 0.2567 & 0.2010 & 0.2553 & 0.1983 \\
	ETTm1 (Window 336) & \textbf{0.0289} & \textbf{0.0216} & 0.2554 & 0.1976 & 0.2559 & 0.1987 & 0.2553 & 0.1981 \\
	ETTm1 (Window 720) & \textbf{0.0283} & \textbf{0.0211} & 0.2744 & 0.2150 & 0.2561 & 0.1997 & 0.2552 & 0.1976 \\
	\midrule
	ETTm2 (Window 96)  & \textbf{0.0497} & \textbf{0.0347} & 0.6334 & 0.4962 & 0.5474 & 0.4176 & 0.5449 & 0.4176 \\
	ETTm2 (Window 120) & \textbf{0.0571} & \textbf{0.0363} & 0.6097 & 0.4683 & 0.5534 & 0.4229 & 0.5553 & 0.4292 \\
	ETTm2 (Window 336) & \textbf{0.0354} & \textbf{0.0162} & 0.6121 & 0.4844 & 0.5473 & 0.4177 & 0.5555 & 0.4267 \\
	ETTm2 (Window 720) & \textbf{0.0689} & \textbf{0.0470} & 0.5472 & 0.4202 & 0.5475 & 0.4195 & 0.6045 & 0.4761 \\
\end{longtable}

\noindent

The experimental results presented in Tables~\ref{tab:ett_results_1} and \ref{tab:ett_results_2} offer a comprehensive evaluation of the proposed model relative to established methods, including N-BEATS, N-HiTS, and TCN, on the ETT datasets. Table~\ref{tab:ett_results_1} reports performance under the non-window setting, where the proposed approach consistently achieves lower error metrics (MSE, RMSE, and MAE) and higher $R^2$ values across all datasets (ETTh1, ETTh2, ETTm1, and ETTm2). These improvements suggest that modeling continuous time series data without partitioning into fixed windows enables the model to capture intrinsic temporal dynamics more effectively, leading to enhanced forecasting accuracy.

In contrast, Table~\ref{tab:ett_results_2} presents the multivariate long-term prediction performance for various fixed window sizes (96, 120, 336, and 720). Even in the window-based setting, the proposed model outperforms the comparative methods, as indicated by the consistently lower RMSE and MAE scores (with the best results highlighted in bold). This demonstrates that the model not only excels in continuous sequence modeling but also scales effectively when applied to windowed data, thereby proving its versatility for both short-term and long-term forecasting tasks.

\begin{figure}[htbp]
	\centering
	\includegraphics[width=1\textwidth,height=8cm]{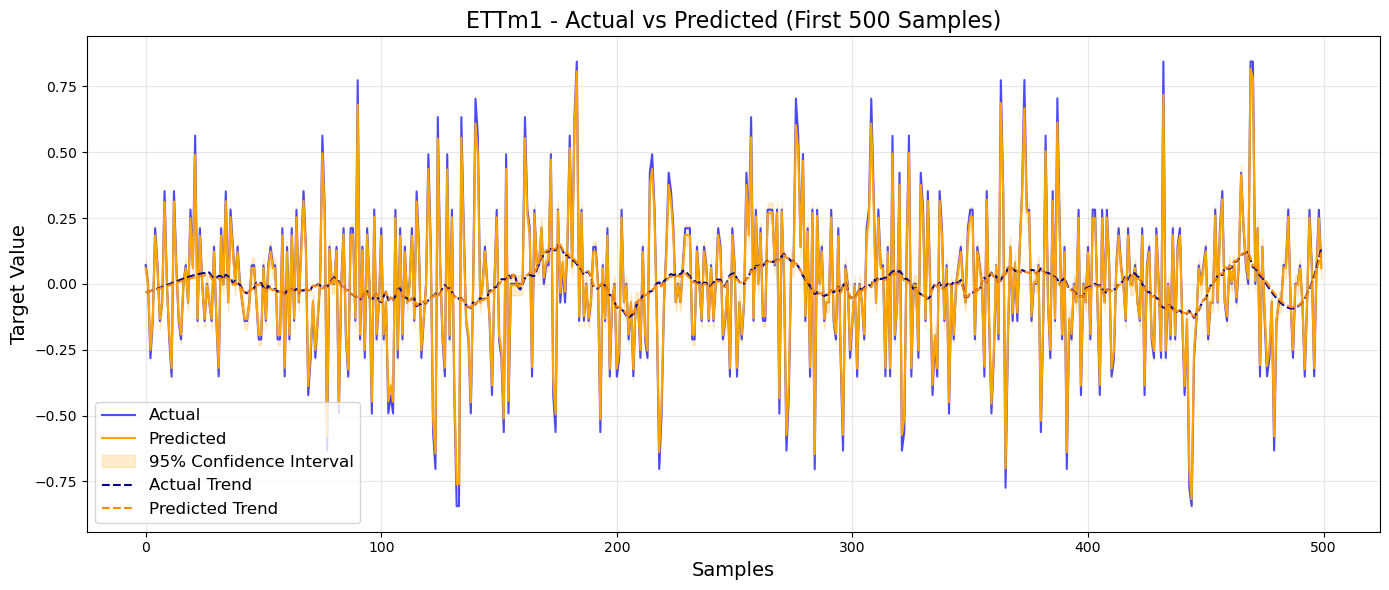}
	\caption{Actual vs. Predicted Values for the ETTm1 Dataset (First 500 Samples). The blue line represents the actual values, the orange line represents the predicted values, and the yellow shaded area indicates the 95\% confidence interval.}
	\label{fig:ettm1_actual_predicted}
\end{figure}

To further illustrate this performance, Figure~\ref{fig:ettm1_actual_predicted} provides a visual comparison of actual versus predicted values for the ETTm1 dataset over the first 500 samples. The close alignment between the actual (blue line) and predicted (orange line) values, coupled with a narrow 95\% confidence interval (yellow shaded area), visually confirms the model's high accuracy and reliability, as reflected by the low MSE of 0.0007 and high $R^2$ of 0.9904 in Table~\ref{tab:ett_results_1}.

\begin{figure}[htbp]
	\centering
	\includegraphics[width=1\textwidth,height=8cm]{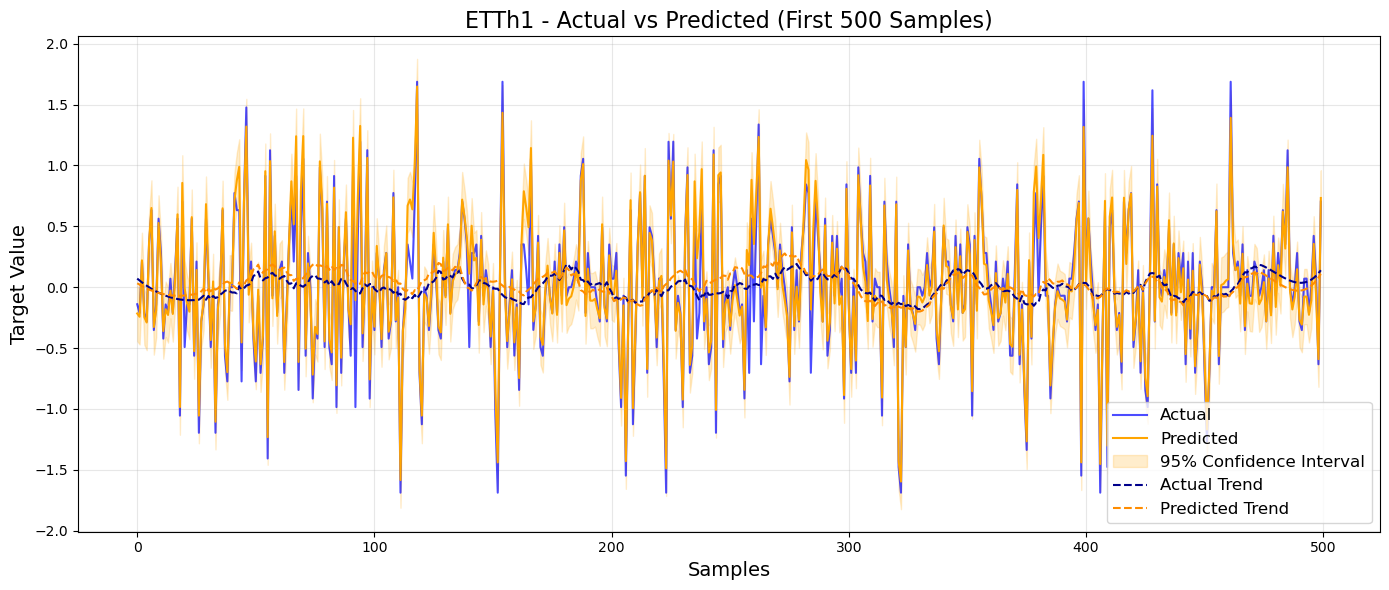}
	\caption{Actual vs. Predicted Values for the ETTh1 Dataset (First 500 Samples). The orange bars represent the actual values, the black dashed line represents the predicted values, and the light orange shaded area indicates the 95\% confidence interval.}
	\label{fig:etth1_actual_predicted}
\end{figure}

Similarly, Figure~\ref{fig:etth1_actual_predicted} shows the actual versus predicted values for the ETTh1 dataset. The predicted values (black dashed line) closely track the actual values (orange bars), with the 95\% confidence interval (light orange shaded area) indicating high prediction certainty, supporting the competitive metrics (e.g., RMSE of 0.1161) in Table~\ref{tab:ett_results_1}.

\begin{figure}[htbp]
	\centering
	\includegraphics[width=1\textwidth,height=8cm]{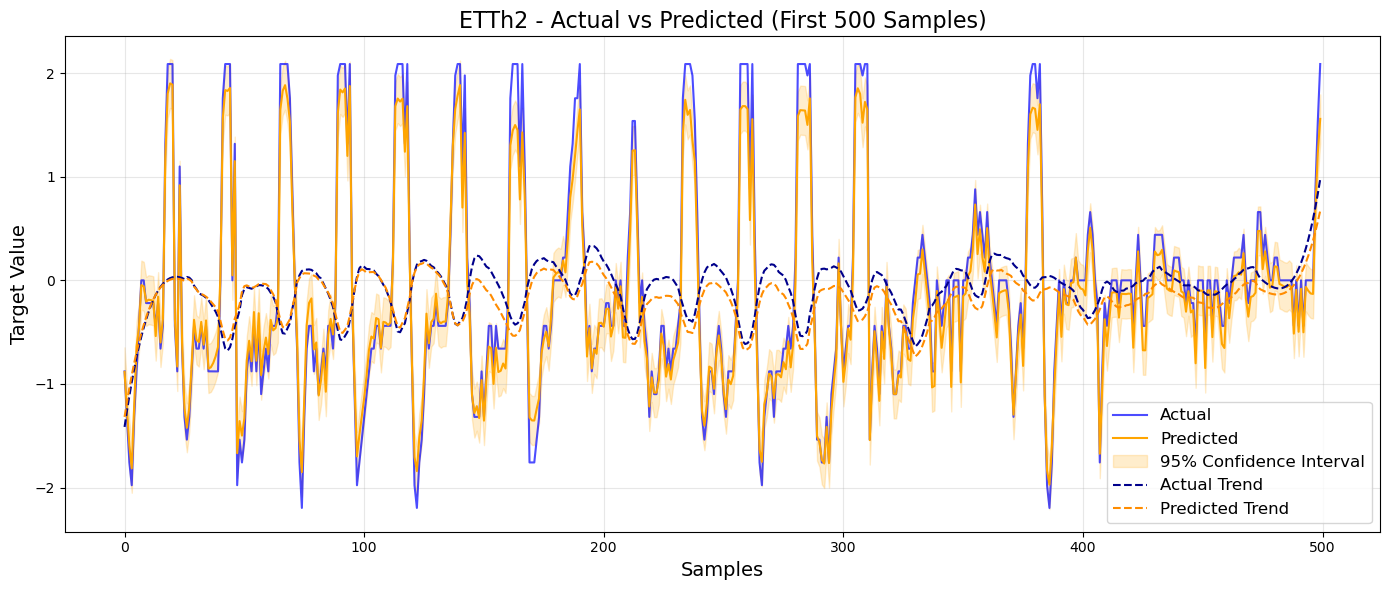}
	\caption{Actual vs. Predicted Values for the ETTh2 Dataset (First 500 Samples). The blue line represents the actual values, and the orange shaded area represents the predicted values with confidence intervals.}
	\label{fig:etth2_actual_predicted}
\end{figure}

For the ETTh2 dataset, Figure~\ref{fig:etth2_actual_predicted} demonstrates the model's ability to capture oscillatory patterns. The predicted values (orange shaded area) align well with the actual values (blue line), reinforcing the low error metrics (e.g., MAE of 0.1076) and high $R^2$ of 0.9825 reported in Table~\ref{tab:ett_results_1}.

\begin{figure}[htbp]
	\centering
	\includegraphics[width=1\textwidth,height=8cm]{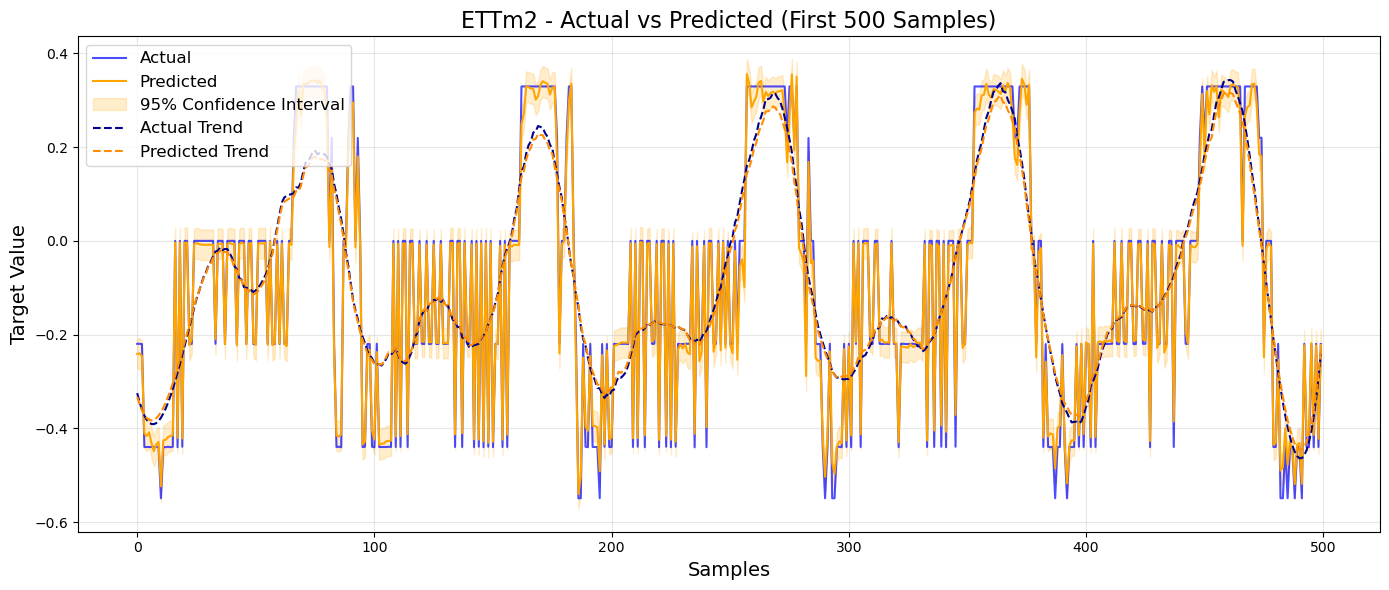}
	\caption{Actual vs. Predicted Values for the ETTm2 Dataset (First 500 Samples). The blue line represents the actual values, the orange line represents the predicted values, and the yellow shaded area indicates the 95\% confidence interval.}
	\label{fig:ettm2_actual_predicted}
\end{figure}

Lastly, Figure~\ref{fig:ettm2_actual_predicted} for the ETTm2 dataset shows a tight fit between actual and predicted values, with the predicted line (orange) closely following the actual line (blue). This visual evidence corroborates the exceptional performance metrics (e.g., MSE of 0.0004, $R^2$ of 0.9941) in Table~\ref{tab:ett_results_1}.

\subsubsection{Renewable Energy Datasets}
\begin{table}[htbp]
	\setlength{\tabcolsep}{4pt}
	\small
	\centering
	\caption{Final Model Comparison Across Waste, Hydro, and Solar Datasets}
	\label{tab:renewable_data}
	\begin{tabular}{|l|l|c|c|c|c|}
		\hline
		\textbf{Dataset} & \textbf{Model} & \textbf{MSE} & \textbf{RMSE} & \textbf{R²} & \textbf{MAE} \\
		\hline
		\multirow{9}{*}{Waste} 
		& Proposed Model & \textbf{0.2732} & \textbf{0.5227} & \textbf{0.9822} & \textbf{0.3709} \\ \cline{2-6}
		& Informer & 0.5501 & 0.7071 & 0.9674 & 0.5500 \\ \cline{2-6}
		& XGBoost      & 1.0785       & 1.0385       & 0.9296       & 0.8150 \\ \cline{2-6}
		& RandomForest & 1.7280       & 1.3145       & 0.8873       & 1.0134 \\ \cline{2-6}
		& Attention    & 151.6369     & 12.3141      & -8.8942      & 11.8709 \\ \cline{2-6}
		& LSTM-Attn    & 2.2135       & 1.4878       & 0.8556       & 1.1120 \\ \cline{2-6}
		& N-BEATS      & 54.8582      & 7.4066       & -2.5794      & 6.3329 \\ \cline{2-6}
		& TCN          & 93.4972      & 9.6694       & -5.1006      & 9.0121 \\ \cline{2-6}
		& RNN\_GRU     & 21.6667      & 4.6547       & -0.4137      & 3.8401 \\
		\hline
		\multirow{9}{*}{Hydro} 
		& Proposed Model & \textbf{1.8470} & \textbf{1.3591} & \textbf{0.9887} & \textbf{1.0205} \\ \cline{2-6}
		& Informer & 3.0419 & 1.7321 & 0.9816 & 1.3300 \\ \cline{2-6}
		& XGBoost      & 6.8049       & 2.6086       & 0.9584       & 1.9523 \\ \cline{2-6}
		& RandomForest & 14.2945      & 3.7808       & 0.9126       & 2.9495 \\ \cline{2-6}
		& Attention    & 173.4518     & 13.1701      & -0.0606      & 10.8026 \\ \cline{2-6}
		& LSTM-Attn    & 3.4334       & 1.8529       & 0.9790       & 1.3491 \\ \cline{2-6}
		& N-BEATS      & 181.6472     & 13.4777      & -0.1107      & 11.0032 \\ \cline{2-6}
		& TCN          & 218.0100     & 14.7652      & -0.3331      & 12.2972 \\ \cline{2-6}
		& RNN\_GRU     & 198.2003     & 14.0784      & -0.2120      & 11.1843 \\
		\hline
		\multirow{9}{*}{Solar} 
		& Proposed Model & \textbf{61.4217} & 7.8372  & 0.8716 & 5.4902 \\ \cline{2-6}
		& Informer & 70.9310 & 8.3676 & 0.8578 & 6.2208 \\ \cline{2-6}
		& XGBoost      & 1166.4377    & 34.1532      & -1.4384      & 26.6778 \\ \cline{2-6}
		& RandomForest & 1208.3395    & 34.7612      & -1.5260      & 27.2310 \\ \cline{2-6}
		& Attention    & 53.2669      & \textbf{7.2984}  & \textbf{0.8886}  & \textbf{5.4474} \\ \cline{2-6}
		& LSTM-Attn    & 1601.8898    & 40.0236      & -2.3487      & 34.0316 \\ \cline{2-6}
		& N-BEATS      & 1453.3674    & 38.1231      & -2.0383      & 29.7683 \\ \cline{2-6}
		& TCN          & 356.1334     & 18.8715      & 0.2555       & 12.5749 \\ \cline{2-6}
		& RNN\_GRU     & 1479.7579    & 38.4676      & -2.0934      & 31.5616 \\
		\hline
	\end{tabular}
\end{table}

The results in Table~\ref{tab:renewable_data} provide a detailed comparison of the final model performance across Waste, Hydro, and Solar renewable energy datasets. The proposed model consistently achieves the best performance in the Waste and Hydro categories, as evidenced by the lowest MSE, RMSE, and MAE values, along with the highest $R^2$ scores. These metrics indicate that the proposed approach is highly effective at capturing the underlying patterns and producing accurate predictions in these domains.

\begin{figure}[htbp]
	\centering
	\includegraphics[width=1\textwidth,height=8cm]{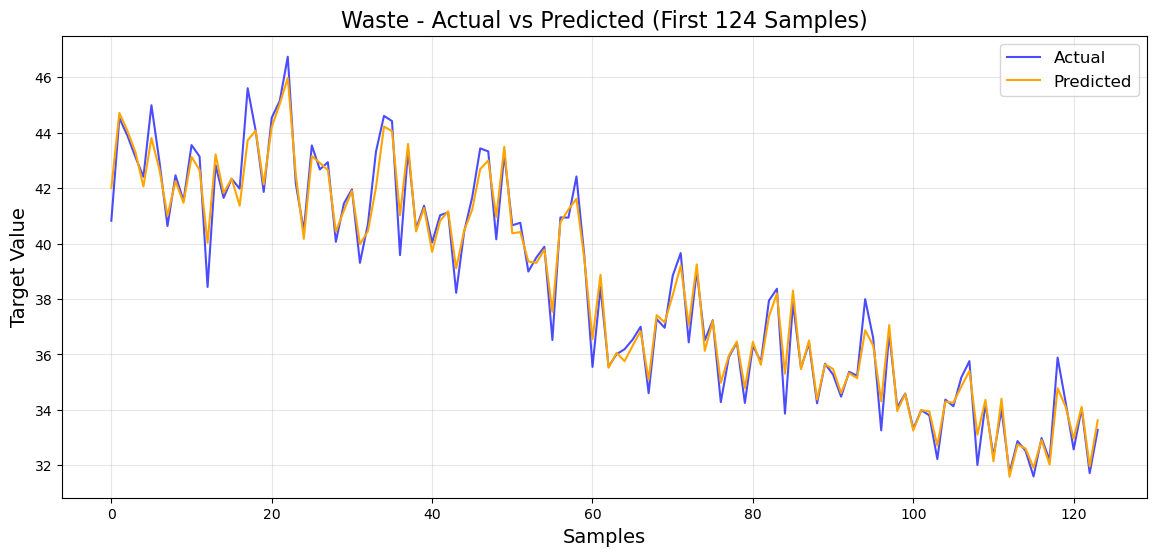}
	\caption{Actual vs. Predicted Values for the Waste Dataset (First 124 Samples). The blue line represents the actual values, and the orange line represents the predicted values.}
	\label{fig:waste_actual_predicted}
\end{figure}

In the Waste dataset, the proposed model achieves an MSE of 0.2732, RMSE of 0.5227, and an $R^2$ value of 0.9822, substantially outperforming traditional machine learning methods such as XGBoost and RandomForest, as well as deep learning models like LSTM-Attn, N-BEATS, TCN, RNN\_GRU and Informer. The dramatic difference in error metrics underlines the effectiveness of the proposed model's design for this type of data. Figure~\ref{fig:waste_actual_predicted} visually supports this, showing the predicted values (orange line) closely tracking the actual values (blue line) over the first 124 samples.

\begin{figure}[htbp]
	\centering
	\includegraphics[width=1\textwidth,height=8cm]{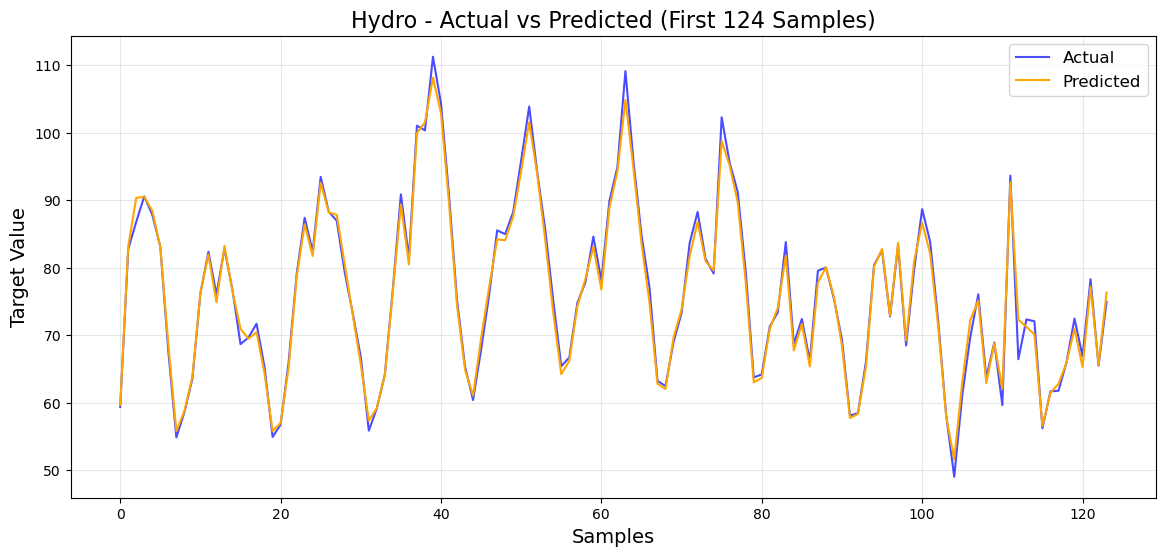}
	\caption{Actual vs. Predicted Values for the Hydro Dataset (First 124 Samples). The blue line represents the actual values, and the orange line represents the predicted values.}
	\label{fig:hydro_actual_predicted}
\end{figure}

A similar trend is observed in the Hydro dataset, where the proposed model records an MSE of 1.8470, RMSE of 1.3591, and an $R^2$ value of 0.9887. This performance surpasses that of other approaches, demonstrating the model's strong capability in handling the variability inherent in hydro energy data. As illustrated in Figure~\ref{fig:hydro_actual_predicted}, the predicted values (orange line) align closely with the actual values (blue line) over the first 124 samples, capturing the overall trend despite some smoothing of short-term fluctuations.

\begin{figure}[htbp]
	\centering
	\includegraphics[width=1\textwidth,height=8cm]{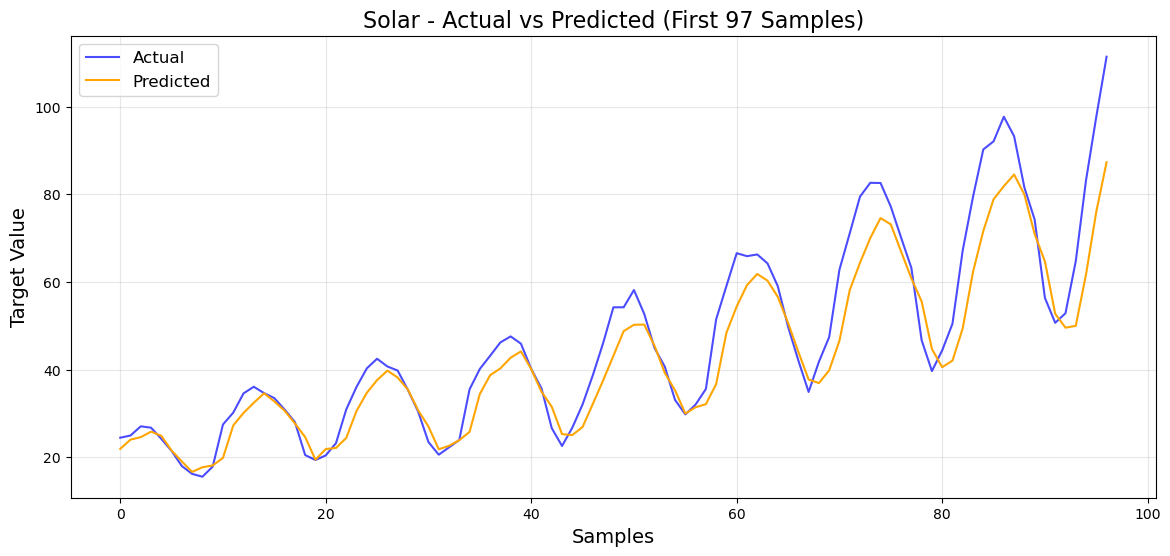}
	\caption{Actual vs. Predicted Values for the Solar Dataset (First 97 Samples). The blue line represents the actual values, and the orange line represents the predicted values.}
	\label{fig:solar_actual_predicted}
\end{figure}

In the Solar dataset, while the proposed model delivers competitive results with an MSE of 61.4217, RMSE of 7.8372, and an $R^2$ of 0.8716, the Attention-based model achieves slightly better performance in terms of RMSE, $R^2$, and MAE. This suggests that for solar data, where different factors might influence the energy output, the Attention mechanism could capture some specific nuances more effectively. Nonetheless, the proposed model maintains a strong overall performance, as shown in Figure~\ref{fig:solar_actual_predicted}, where the predicted values (orange line) generally follow the actual values (blue line) over the first 97 samples, with some deviations in peak amplitudes.

To provide a comprehensive summary of the proposed model's performance across all evaluated datasets, Table~\ref{tab:summary_results} has been presented, which captures the key highlights by showing the number of times the proposed model achieves the best results for each metric compared to the baseline models. This summary aggregates results from both the ETT datasets (including the non-window setting and various window sizes) and the renewable energy datasets (Waste, Hydro, and Solar). As indicated in the table, the proposed model consistently outperforms the baselines in the majority of settings. Specifically, it achieves the best performance in 22 out of 23 settings for both RMSE and MAE metrics, and in 6 out of 7 settings for MSE and $R^2$ metrics. This demonstrates the model's robustness and effectiveness across diverse time series forecasting tasks.
\begin{table}[htbp]
	\centering
	\caption{Summary of Best Model Results Across All Datasets}
	\label{tab:summary_results}
	\begin{tabular}{|l|c|c|}
		\hline
		Metric & Total settings & Wins by proposed model \\
		\hline
		MSE    & 7  & 6 \\ \hline
		RMSE   & 23 & 22 \\ \hline
		MAE    & 23 & 22 \\ \hline
		$R^2$  & 7  & 6 \\ 
		\hline
	\end{tabular}
\end{table}

\subsubsection{Execution Time and Optimal Hyperparameters}

In addition to predictive performance, the efficiency of the model in terms of training and testing times is crucial for practical applications. This subsection presents the execution times and the optimal hyperparameters identified through the training process for both non-windowed and multi-window settings across various datasets. The hyperparameters include the hidden dimension (\texttt{hidden\_dim}), batch size (\texttt{batch\_size}), number of epochs (\texttt{epochs}), dropout rate (\texttt{dropout\_rate}), and learning rate (\texttt{learning\_rate}). These parameters were tuned to achieve the best performance for each dataset and setting.

\begin{table}[htbp]
	\centering
	\caption{Execution Times and Optimal Hyperparameters (Hidden Dimension, Batch Size, Epochs, Dropout Rate, Learning Rate) for Non-Windowed and Multi-Window Settings Across Datasets}
	\label{tab:execution_times}
	\begin{tabular}{|l|l|c|c|c|}
		\hline
		\textbf{Dataset} & \textbf{Setting} & \textbf{Best Parameters} & \textbf{Training Time (s)} & \textbf{Testing Time (s)} \\
		\hline
		ETTh1 & Non-windowed & 64, 16, 100, 0.5, 0.000307 & 182.91 & 0.83 \\
		ETTh2 & Non-windowed & 128, 32, 30, 0.1, 0.005217 & 112.82 & 0.94 \\
		ETTm1 & Non-windowed & 48, 128, 30, 0.5, 0.005163 & 109.62 & 1.80 \\
		ETTm2 & Non-windowed & 48, 32, 30, 0.1, 0.000515 & 333.38 & 1.72 \\
		Waste & Non-windowed & 446, 16, 207, 0.2, 0.000238 & 64.03 & 0.56 \\
		Hydro & Non-windowed & 446, 16, 207, 0.2, 0.000238 & 56.87 & 1.12 \\
		Solar & Non-windowed & 96, 8, 63, 0.0001 & 15.62 & 0.60 \\
		\hline
		ETTh1 & Window 96 &  96 16, 90, 0.2, 0.002810 & 55.97 & 3.79 \\
		ETTh1 & Window 120 & 48, 64, 100, 0.1, 0.002016 & 46.48 & 3.33 \\
		ETTh1 & Window 336 & 128, 64, 50, 0.1, 0.000646 & 39.59 & 3.37 \\
		ETTh1 & Window 720 & 112, 16, 50, 0.5, 0.000478 & 96.52 & 3.34 \\
		ETTh2 & Window 96 & 128, 128, 50, 0.4, 0.000171 & 39.90 & 3.34 \\
		ETTh2 & Window 120 & 112, 16, 90, 0.1, 0.000871 & 87.57 & 3.43 \\
		ETTh2 & Window 336 & 128, 32, 90, 0.1, 0.001115 & 49.49 & 3.43 \\
		ETTh2 & Window 720 & 128, 16, 80, 0.1, 0.000176 & 69.07 & 3.37 \\
		ETTm1 & Window 96 & 112, 128, 60, 0.1, 0.000301 & 94.69 & 3.93 \\
		ETTm1 & Window 120 & 112, 32, 50, 0.1, 0.003735 & 98.09 & 3.92 \\
		ETTm1 & Window 336 & 32, 16, 90, 0.1, 0.003049 & 233.31 & 4.05 \\
		ETTm1 & Window 720 & 128, 16, 30, 0.1, 0.000808 & 228.08 & 4.31 \\
		ETTm2 & Window 96 & 80, 32, 90, 0.2, 0.000980 & 124.49 & 4.27 \\
		ETTm2 & Window 120 & 128, 128, 50, 0.3, 0.001495 & 62.39 & 4.24 \\
		ETTm2 & Window 336 & 128, 16, 80, 0.1, 0.000172 & 450.05 & 4.35 \\
		ETTm2 & Window 720 & 32, 128, 30, 0.1, 0.002456 & 62.05 & 4.35 \\
		\hline
	\end{tabular}

\end{table}

The execution times vary significantly across different datasets and settings. In the non-windowed setting, training times range from 15.62 seconds for the Solar dataset to 333.38 seconds for ETTm2, reflecting differences in dataset size and model complexity. Testing times are generally much shorter, ranging from 0.56 seconds to 1.80 seconds. For multi-window settings, training times tend to be shorter than in the non-windowed setting for some datasets, likely due to reduced sequence lengths. For instance, for ETTh1, training times decrease from 182.91 seconds in the non-windowed setting to between 39.59 and 96.52 seconds across various window sizes. However, testing times are slightly higher in the multi-window settings, possibly due to the need to process multiple windows. The optimal hyperparameters also vary across datasets and settings, indicating the importance of tuning for specific tasks. Notably, larger hidden dimensions and more epochs are used for datasets like Waste and Hydro, which may require greater capacity to model complex patterns.

\subsubsection{Memory Consumption}

\begin{table}[htbp]
	\small
	\centering
	\setlength{\tabcolsep}{2pt}
	\caption{Memory Consumption for ETT and Energy Datasets (Non-Windowed Setting)}
	\label{tab:memory_consumption}
	\begin{tabular}{|l|c|c|c|}
		
		\hline
		\textbf{Dataset} & \textbf{Data Memory (MB)} & \textbf{Model Parameters Memory (MB)} & \textbf{Total Memory (MB)} \\
		\hline
		ETTh1 & 2.06 & 0.06 & 2.12 \\
		ETTh2 & 2.06 & 0.24 & 2.31 \\
		ETTm1 & 8.24 & 0.04 & 8.28 \\
		ETTm2 & 8.24 & 0.04 & 8.28 \\
		Waste & 0.18 & 2.90 & 3.09 \\
		Hydro & 0.19 & 2.33 & 2.52 \\
		Solar & 0.15 & 3.44 & 3.59 \\
		\hline
	\end{tabular}
\end{table}

The memory consumption analysis for the non-windowed setting reveals efficient resource usage across datasets. From Table~\ref{tab:memory_consumption}, it can be seen that ETTm datasets (ETTm1 and ETTm2) exhibit the highest total memory usage at 8.28 MB, primarily due to their larger data memory (8.24 MB) stemming from higher resolution. In contrast, the Energy datasets (Waste, Hydro, Solar) have lower data memory (0.15--0.19 MB) but higher model parameters memory (2.33--3.44 MB), driven by larger hidden dimensions in their configurations. The Solar dataset consumes the most memory overall (3.59 MB), largely due to its complex model parameters. Batch processing (found 0 MB) and adjacency matrix memory are negligible, indicating an optimized model design suitable for practical deployment.

\section{Comprehensive Error Analysis Across Datasets}
Building on the results presented in the previous section, this subsection provides an in-depth analysis of the prediction errors across all datasets to evaluate the model's predictive performance and robustness. In addition to standard error metrics—Mean Error (ME), Standard Deviation of Errors (SDE), Median Error, Maximum Error, and Minimum Error—the analysis also includes Median Absolute Deviation (MAD), Theil's U Statistic, and the 95\% Coverage Probability. These metrics collectively characterize the central tendency, dispersion, and uncertainty of the model's predictions.

Table~\ref{tab:error_analysis} summarizes these metrics for the ETTh1, ETTh2, ETTm1, ETTm2, Hydro, Waste, and Solar datasets.

\begin{table}[ht]
	\centering
	\caption{Comprehensive Error Metrics Across Datasets}
	\label{tab:error_analysis}
	\begin{tabular}{|l|c|c|c|c|c|c|c|}
		\hline
		\textbf{Dataset} & \textbf{ME} & \textbf{SDE} & \textbf{Median Error} & \textbf{Max Error} & \textbf{Min Error} & \textbf{MAD} & \textbf{Theil's U} \\
		\hline
		ETTh1 & 0.0498  & 0.1416  & 0.0312  & 0.9872  & -0.6823 & 0.0627  & 0.1838 \\ \hline
		ETTh2 & 0.0116  & 0.0923  & -0.0144 & 0.2725  & -0.3753 & 0.0509  & 0.1348 \\ \hline
		ETTm1 & -0.0398 & 0.1129  & -0.0524 & 0.4651  & -0.5893 & 0.0858  & 0.3072 \\ \hline
		ETTm2 & -0.0112 & 0.0268  & -0.0064 & 0.4912  & -0.3360 & 0.0082  & 0.1524 \\ \hline
		Hydro & 0.6772  & 1.2209  & 0.8692  & 3.5335  & -5.9906 & 0.9668  & 0.1400 \\ \hline
		Waste & -0.0116 & 0.5226  & 0.0313  & 1.8772  & -1.5942 & 0.2663  & 0.2365 \\ \hline
		Solar & 3.3500  & 7.2618  & 4.0029  & 20.9364 & -15.7971& 5.0883  & 1.1346 \\ 
		\hline
	\end{tabular}
	
	\vspace{0.5em}
	\begin{tabular}{|l|c|}
		\hline
		\textbf{Dataset} & \textbf{95\% Coverage Probability} \\
		\hline
		ETTh1 & 92.6808\% \\ \hline
		ETTh2 & 93.9437\% \\ \hline
		ETTm1 & 94.9483\% \\ \hline
		ETTm2 & 94.9842\% \\ \hline
		Hydro & 95.1613\% \\ \hline
		Waste & 91.1290\% \\ \hline
		Solar & 89.6907\% \\ 
		\hline
	\end{tabular}
\end{table}

\paragraph{ETTh1 and ETTh2 Datasets:}
For the high-frequency ETTh datasets, the error metrics indicate robust performance. \textbf{ETTh1} exhibits a slight positive bias with an ME of 0.0498 and a narrow dispersion (SDE = 0.1416), accompanied by a median error of 0.0312. The error range is also modest (Max = 0.9872, Min = -0.6823), with a MAD of 0.0627 and a Theil's U of 0.1838, while the prediction intervals cover 92.68\% of true values. Similarly, \textbf{ETTh2} shows an almost negligible bias (ME = 0.0116) and even lower variability (SDE = 0.0923) with a median error of -0.0144. The error limits (Max = 0.2725, Min = -0.3753), MAD of 0.0509, and Theil's U of 0.1348, along with a 93.94\% coverage probability, underline the precise and reliable performance on this dataset.

\paragraph{ETTm1 and ETTm2 Datasets:}
For the medium-frequency ETTm datasets, \textbf{ETTm1} reveals a slight negative bias (ME = -0.0398) and moderate error variability (SDE = 0.1129), with a median error of -0.0524. The error range spans from -0.5893 to 0.4651, with a MAD of 0.0858 and a Theil's U of 0.3072; the prediction intervals cover 94.95\% of the true values. \textbf{ETTm2} demonstrates minimal bias (ME = -0.0112) and exceptionally low variability (SDE = 0.0268), with nearly zero median error (-0.0064) and a narrow error range (Max = 0.4912, Min = -0.3360). Its MAD is particularly low (0.0082) and Theil's U is 0.1524, with a coverage probability of 94.98\%, indicating excellent consistency.

\paragraph{Hydro Dataset:}
The Hydro dataset exhibits a more pronounced positive bias (ME = 0.6772) and higher variability (SDE = 1.2209). The error distribution is wider (Max = 3.5335, Min = -5.9906) with a median error of 0.8692. Despite the larger error spread, the MAD of 0.9668 and a Theil's U of 0.1400 suggest that the model still significantly outperforms a naive benchmark. The prediction intervals capture 95.16\% of true values, demonstrating reliable uncertainty estimation.

\paragraph{Waste Dataset:}
For the Waste dataset, the ME is nearly zero (-0.0116) with an SDE of 0.5226 and a median error of 0.0313, though the error range is broader (Max = 1.8772, Min = -1.5942) compared to the ETTh and ETTm datasets. The MAD is 0.2663 and Theil's U is 0.2365, while the prediction intervals cover 91.13\% of the data, indicating moderate variability but generally unbiased predictions.

\paragraph{Solar Dataset:}
The Solar dataset is the most challenging, with a high ME of 3.3500 and substantial variability (SDE = 7.2618). The errors are widely dispersed (Max = 20.9364, Min = -15.7971) and the median error is 4.0029. With a MAD of 5.0883 and a Theil's U of 1.1346, the model's performance is close to or even below that of a naive forecast. The 95\% Coverage Probability is the lowest at 89.69\%, indicating less reliable prediction intervals.

\paragraph{Comparison and Insights:}
The error analysis across these datasets illustrates the model's versatility and areas for potential improvement. The ETTh1, ETTh2, ETTm1, and ETTm2 datasets consistently exhibit minimal bias and low variability, reflecting strong performance on regular time-series data. In contrast, the Hydro and Waste datasets display higher error dispersion—with Hydro showing a noticeable positive bias—and the Solar dataset highlights significant challenges with high bias and variability. These comprehensive error metrics provide a nuanced view of the model’s predictive accuracy and suggest directions for further refinement, particularly for complex datasets such as Solar.

\section{Ablation Study}
An ablation study was conducted to assess the contributions of individual components within the proposed model, achieved through the systematic removal of each pathway. The evaluation utilized the ETTh1 dataset, with results summarized in Table~\ref{tab:ablation_part1}, which presents key performance and computational metrics, supplemented by Theil's U Statistic and 95\% Coverage Probability in Table~\ref{tab:ablation_part2}. Together, these metrics provide a robust framework for evaluating the model's efficacy and the relative importance of its components.

\begin{table}[ht]
	\centering
	\caption{Ablation Study Results on ETTh1 Dataset (Part 1: Core Metrics)}
	\label{tab:ablation_part1}
	\begin{tabular}{|l|c|c|c|c|c|c|c|c|}
		\hline
		\textbf{Configuration} & \textbf{$R^2$} & \textbf{MAE} & \textbf{MSE} & \textbf{RMSE} & \textbf{ME} & \textbf{SDE} & \textbf{Train (s)} & \textbf{Test (s)} \\
		\hline
		Full Model             & 0.9606 & 0.0843 & 0.0135 & 0.1161 & 0.0498 & 0.1416 & 84.76 & 0.82 \\ 
		No ODE                & 0.9510 & 0.0925 & 0.0143 & 0.1155 & 0.0500 & 0.1541 & 21.18 & 0.18 \\ 
		No Daubechies         & 0.9522 & 0.0921 & 0.0172 & 0.1059 & 0.0591 & 0.1949 & 121.15 & 0.76 \\ 
		No Parametric         & 0.9526 & 0.1008 & 0.0162 & 0.1273 & 0.0791 & 0.1797 & 116.00 & 0.81 \\ 
		No Dense              & 0.9500 & 0.0953 & 0.0268 & 0.1827 & 0.0672 & 0.2824 & 84.31 & 1.08 \\ 
		Only ODE              & 0.7447 & 0.1912 & 0.0874 & 0.2956 & -0.0535 & 0.2907 & 105.21 & 0.81 \\

		\hline
	\end{tabular}
\end{table}

\begin{table}[ht]
	\centering
	\caption{Ablation Study Results on ETTh1 Dataset (Part 2: Additional Metrics)}
	\label{tab:ablation_part2}
	\begin{tabular}{|l|c|c|}
		\hline
		\textbf{Configuration} & \textbf{Theil's U Statistic} & \textbf{95\% Coverage Probability} \\
		\hline
		Full Model             & 0.1838 & 92.6808 \\ 
		No ODE                & 0.1414 & 92.594719 \\ 
		No Daubechies         & 0.1297 & 93.398393 \\ 
		No Parametric         & 0.1558 & 88.002296 \\ 
		No Dense              & 0.1012 & 94.546498 \\ 
		Only ODE              & 0.2616 & 93.752009 \\ 
		
		\hline
	\end{tabular}
\end{table}

The ablation study elucidated the distinct roles of each model component. Removing the ODE component reduced \(R^2\) to 0.9510 and increased MAE to 0.0925, MSE to 0.0143, and SDE to 0.1541, suggesting a modest decline in accuracy and consistency, though RMSE slightly improved to 0.1155. Notably, this configuration slashed training time to 21.18s and testing time to 0.18s, with Theil's U Statistic dropping to 0.1414---indicating improved forecast accuracy over a naive baseline---while 95\% Coverage Probability fell to 92.594719\%, implying a minor reliability trade-off. Excluding the Daubechies transform yielded an \(R^2\) of 0.9522, a reduced RMSE of 0.1059, and a slight MAE increase to 0.0921, but MSE rose to 0.0172 and SDE to 0.1949, pointing to greater error variability. Training time surged to 121.15s, though Theil's U Statistic improved to 0.1297 and 95\% Coverage Probability increased to 93.398393\%, suggesting enhanced reliability despite higher computational cost. Omitting the parametric transform resulted in an \(R^2\) of 0.9526, with increases in MAE (0.1008), MSE (0.0162), RMSE (0.1273), ME (0.0791), and SDE (0.1797). Training time was 116.00s, with Theil's U Statistic at 0.1558 and a notably lower 95\% Coverage Probability of 88.002296\%, indicating diminished reliability in uncertainty estimates. Removing the dense layers led to an \(R^2\) of 0.9500, with significant increases in MSE (0.0268), RMSE (0.1827), MAE (0.0953), ME (0.0672), and SDE (0.2824). Training time remained stable at 84.31s, but testing time rose to 1.08s. Theil's U Statistic improved to 0.1012---reflecting strong forecast accuracy relative to a naive model---while 95\% Coverage Probability reached 94.546498\%, suggesting robust uncertainty quantification despite degraded performance. Finally, using only the ODE component caused a sharp decline in performance, with \(R^2\) falling to 0.7447, MAE rising to 0.1912, MSE to 0.0874, RMSE to 0.2956, and SDE to 0.2907, alongside a negative ME of -0.0535. Training time was 105.21s, with Theil's U Statistic at 0.2616 and 95\% Coverage Probability at 93.752009\%. This configuration underscores the critical need for the multi-path architecture to achieve robust forecasting. Overall, the ablation study confirms that each component enhances the model's predictive capability, with trade-offs in computational efficiency and reliability varying by pathway. The full model strikes an optimal balance across these dimensions.

The ablation study assessed the model's architectural components, identifying essential elements for performance. However, it did not clarify how input features influence predictions. To address this, SHAP analysis was used to quantify feature contributions, providing deeper insights into the data's impact. These findings can guide model refinement and feature engineering.

\section{SHAP-Based Interpretability of Predictive Models Across Diverse Energy Datasets}

This section elucidates the interpretability of predictive models for energy time series forecasting through SHAP (SHapley Additive exPlanations) \cite{shap} analysis, applied to multiple datasets: ETTh1 (hourly), ETTm1 (minute\_level), waste, hydroelectric, and solar energy consumption. SHAP values were employed to quantify the contributions of selected features to model outputs, offering both global and local perspectives on model behavior. Global interpretability is assessed via bar, summary, and dependence plots, while local insights are derived from waterfall plots and feature contribution tables for Instance 0 \cite{local}. This comprehensive analysis not only highlights the dominant features driving predictions but also underscores the varying influence of temporal and statistical features across datasets, providing actionable insights for refining forecasting methodologies.

\subsection{ETTh1 Dataset Analysis}

\begin{figure}[htbp]
	\centering
	\begin{subfigure}{0.49\textwidth}
		\includegraphics[width=\textwidth,height=7cm]{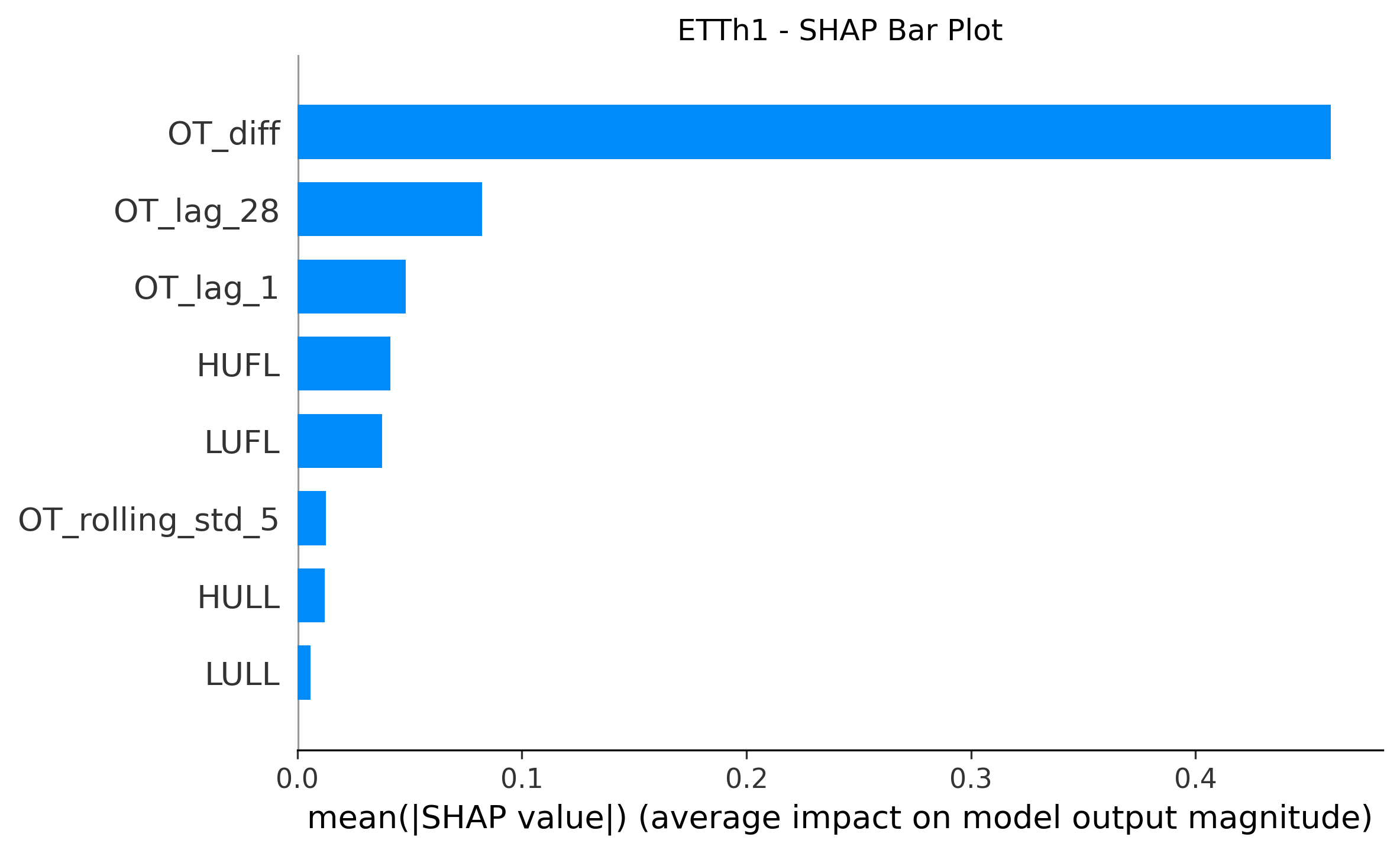}
		\caption{SHAP Bar Plot}
		\label{fig:etth1_shap_bar}
	\end{subfigure}
	\hfill
	\begin{subfigure}{0.49\textwidth}
		\includegraphics[width=\textwidth,height=7cm]{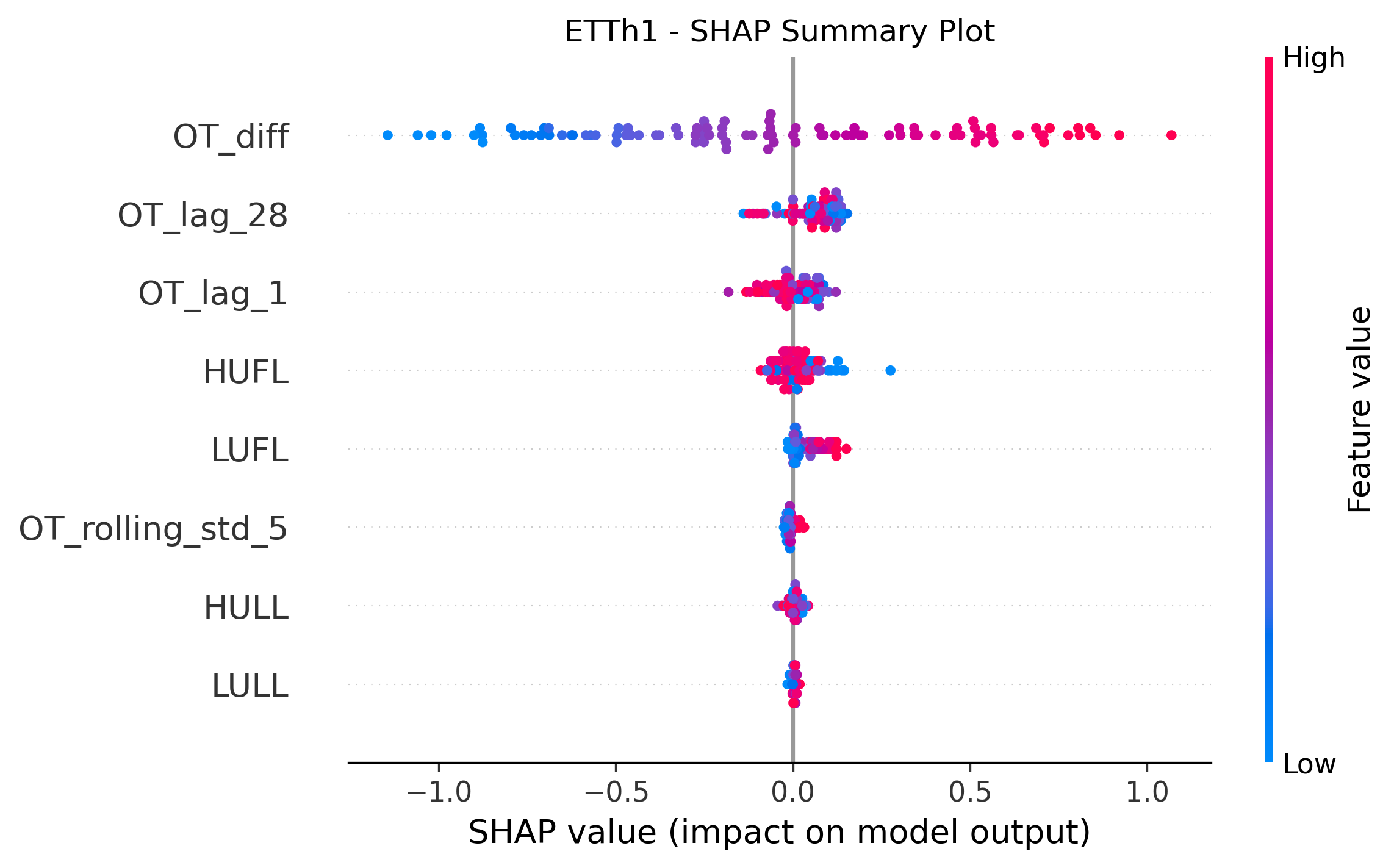}
		\caption{SHAP Summary Plot}
		\label{fig:etth1_shap_summary}
	\end{subfigure}
	\caption{ETTh1 - Global SHAP Visualizations}
\end{figure}

\begin{figure}[htbp]
	\centering
	\includegraphics[width=0.8\textwidth,height=7cm]{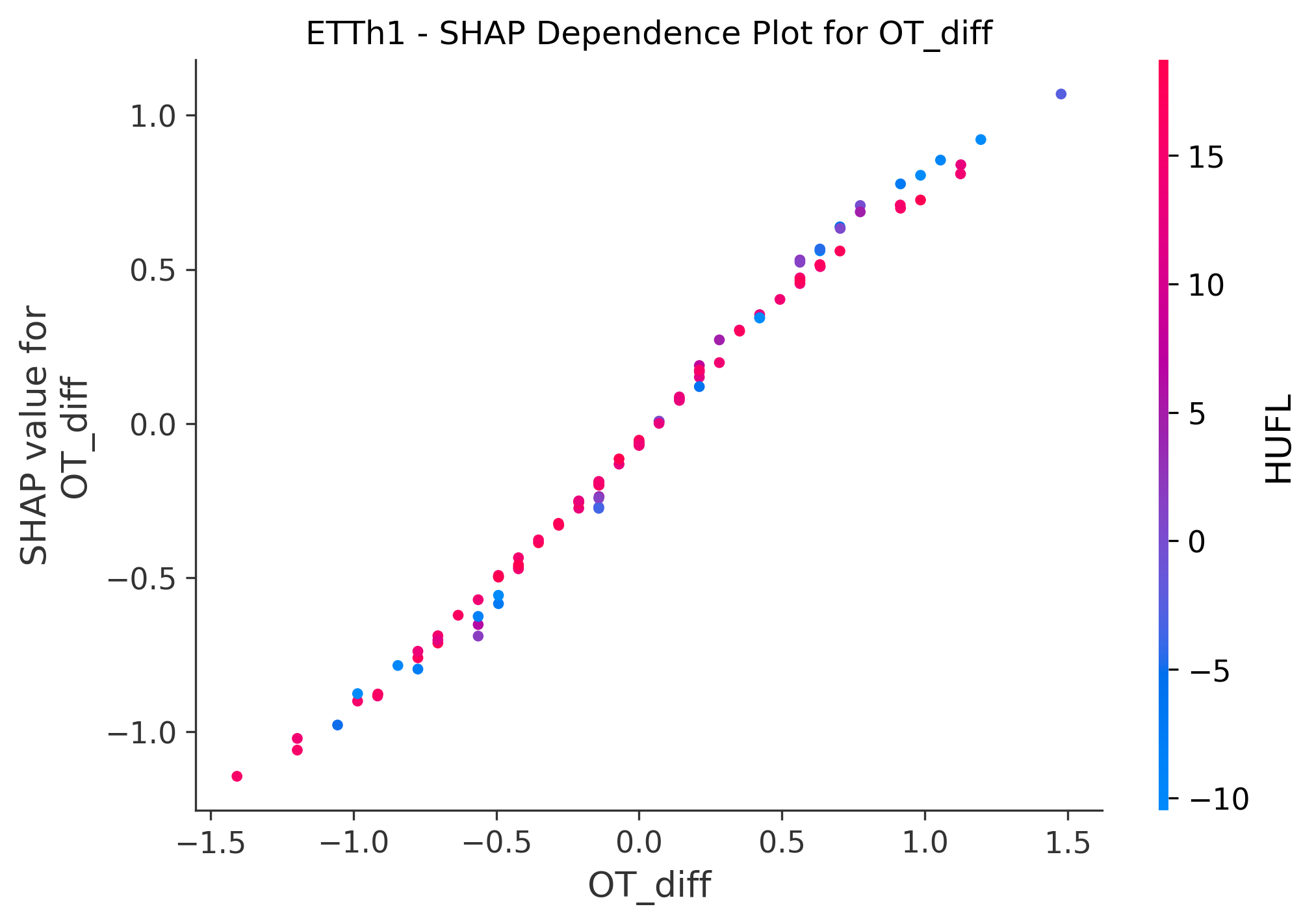}
	\caption{ETTh1 - SHAP Dependence Plot for OT\_diff}
	\label{fig:etth1_shap_dependence}
\end{figure}

\begin{figure}[htbp]
	\centering
	\includegraphics[width=0.8\textwidth]{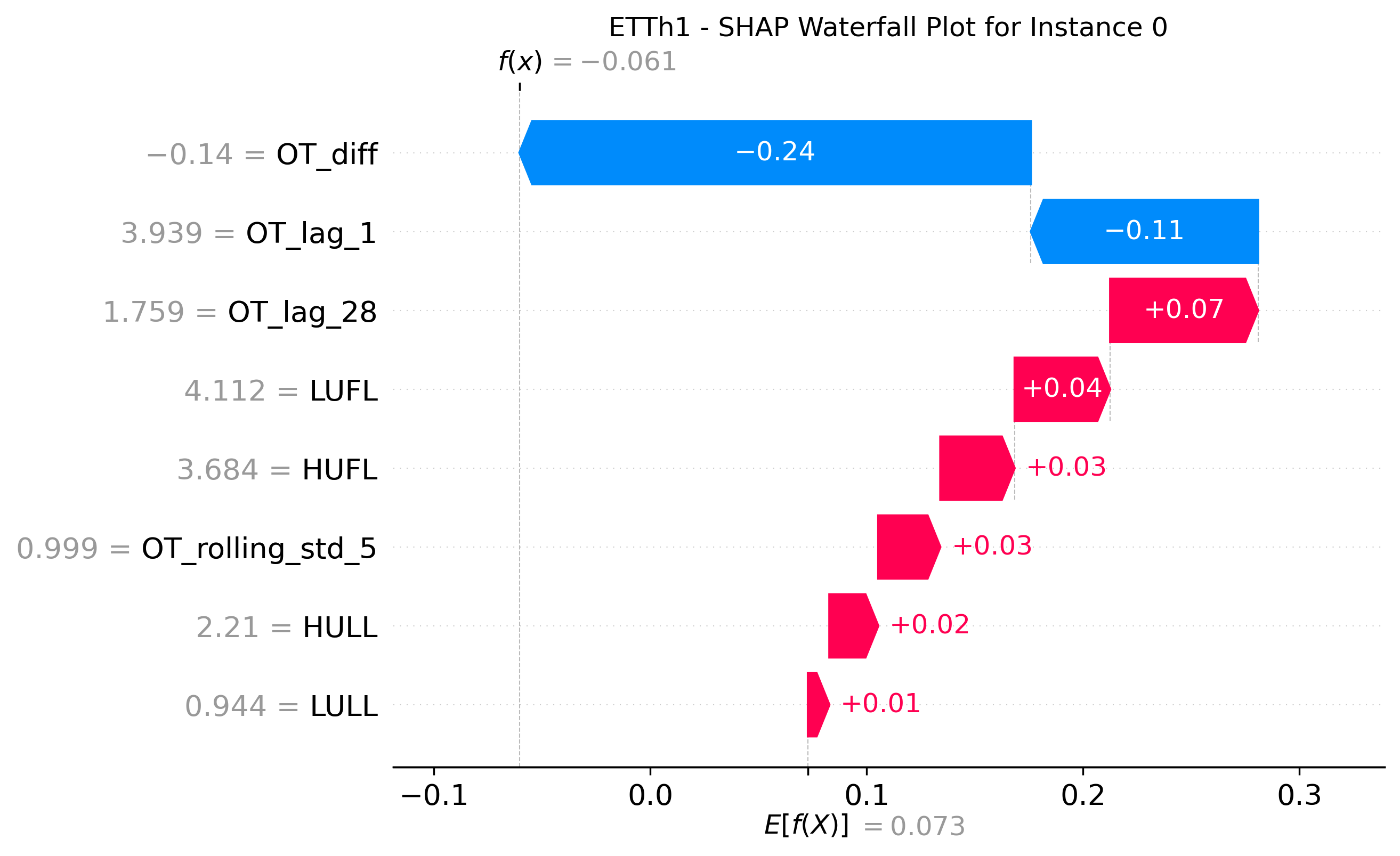}
	\caption{ETTh1 - SHAP Waterfall Plot for Instance 0}
	\label{fig:etth1_shap_waterfall}
\end{figure}

The SHAP analysis for the ETTh1 dataset, which captures hourly energy data, identifies OT\_diff as the most influential feature, as evidenced by the bar plot (Figure~\ref{fig:etth1_shap_bar}) showing its substantial average impact on model output, followed by OT\_lag\_28 and OT\_lag\_1. The summary plot (Figure~\ref{fig:etth1_shap_summary}) reveals that higher OT\_diff values generally reduce the predicted output, while lag features exhibit mixed effects depending on their magnitude. A dependence plot (Figure~\ref{fig:etth1_shap_dependence}) further illustrates a positive correlation between OT\_diff and its SHAP values, with an interaction effect from HUFL modulating this relationship. At the local level, the waterfall plot for Instance 0 (Figure~\ref{fig:etth1_shap_waterfall}) demonstrates a negative contribution from OT\_diff (-0.24) to the prediction (\(f(x) = -0.061\)), offset by contributions from OT\_lag\_28 (+0.07) and OT\_lag\_1 (-0.11). These local contributions are detailed in Table~\ref{tab:etth1_contributions}, emphasizing the pivotal role of differential and lag features in hourly energy predictions.

\begin{table}[htbp]
	\centering
	\caption{Local Feature Contributions for Instance 0 (ETTh1)}
	\label{tab:etth1_contributions}
	\begin{tabular}{|l|c|c|c|}
		\hline
		Feature & SHAP Value & Feature Value & Contribution Impact \\
		\hline
		OT\_diff & -0.236445 & -0.140000 & 0.033102 \\
		OT\_lag\_1 & -0.105012 & 3.939000 & -0.413642 \\
		OT\_lag\_28 & 0.068362 & 1.759000 & 0.120248 \\
		LUFL & 0.044168 & 4.112000 & 0.181620 \\
		HUFL & 0.034347 & 3.684000 & 0.126534 \\
		OT\_rolling\_std\_5 & 0.028701 & 0.999161 & 0.028677 \\
		HULL & 0.022598 & 2.210000 & 0.049942 \\
		LULL & 0.009909 & 0.944000 & 0.009354 \\
		\hline
	\end{tabular}
\end{table}

\subsection{ETTm1 Dataset Analysis}

\begin{figure}[htbp]
	\centering
	\begin{subfigure}{0.49\textwidth}
		\includegraphics[width=\textwidth,height=7cm]{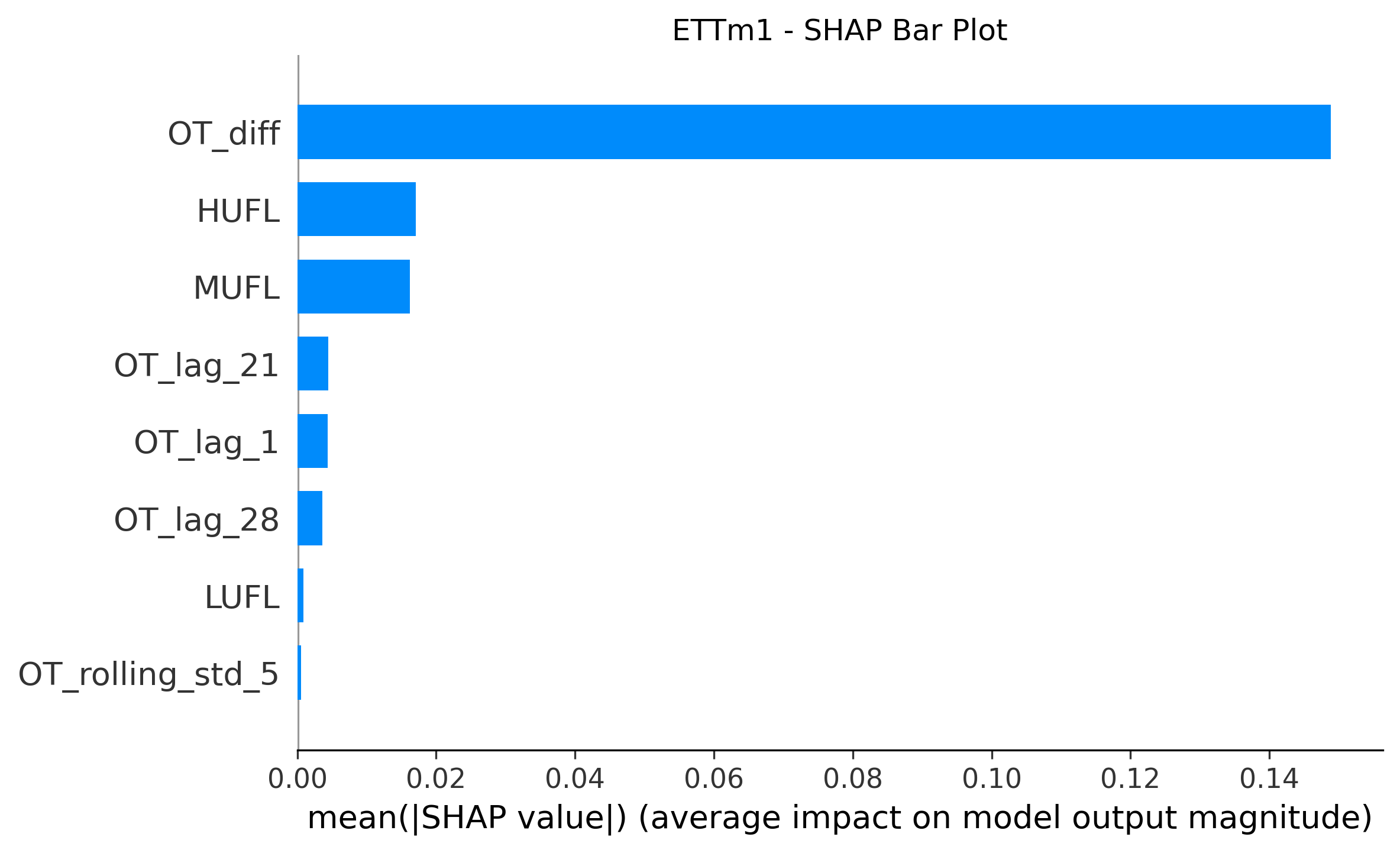}
		\caption{SHAP Bar Plot}
		\label{fig:ettm1_shap_bar}
	\end{subfigure}
	\hfill
	\begin{subfigure}{0.49\textwidth}
		\includegraphics[width=\textwidth,height=7cm]{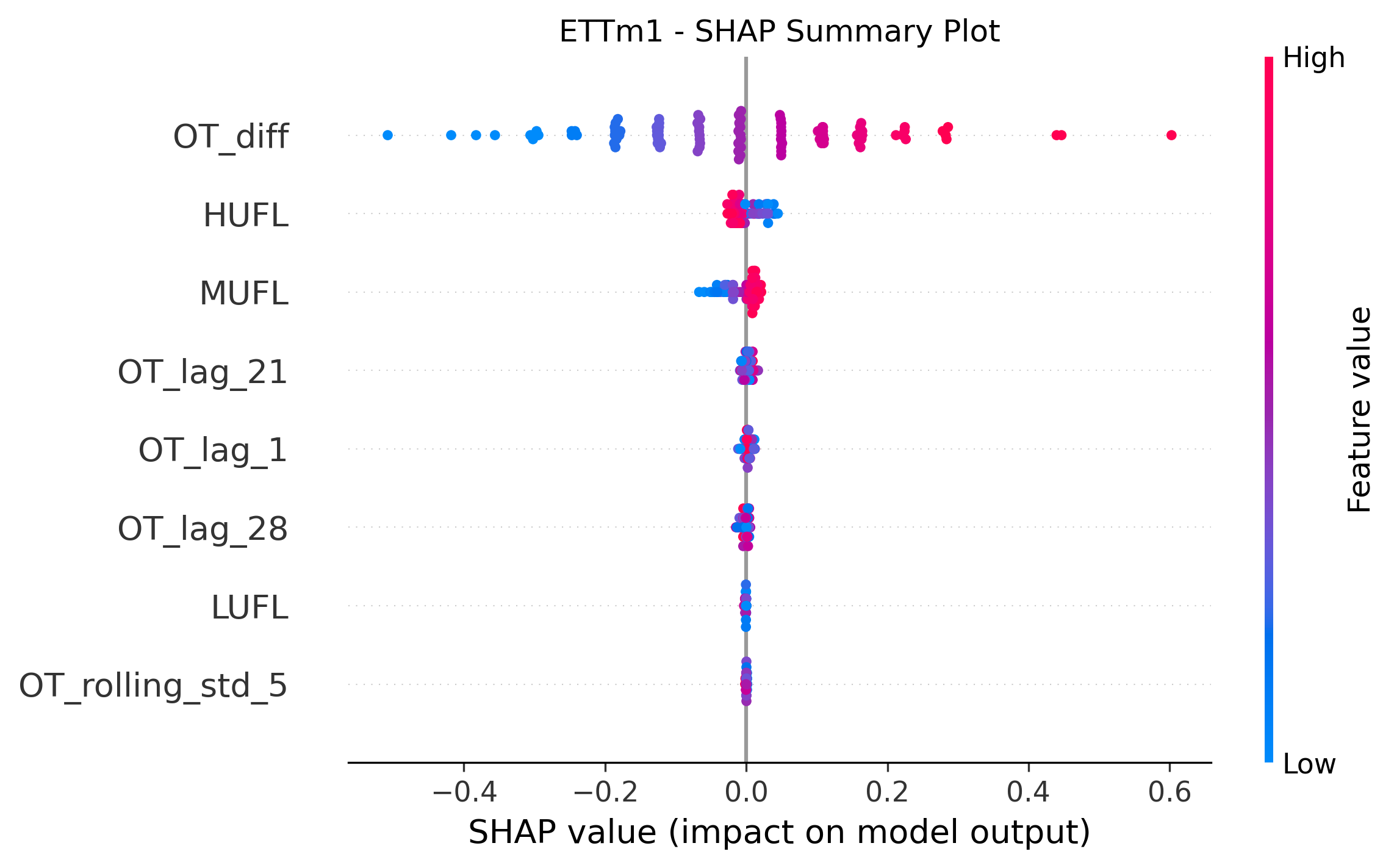}
		\caption{SHAP Summary Plot}
		\label{fig:ettm1_shap_summary}
	\end{subfigure}
	\caption{ETTm1 - Global SHAP Visualizations}
\end{figure}

\begin{figure}[htbp]
	\centering
	\includegraphics[width=0.8\textwidth,height=7cm]{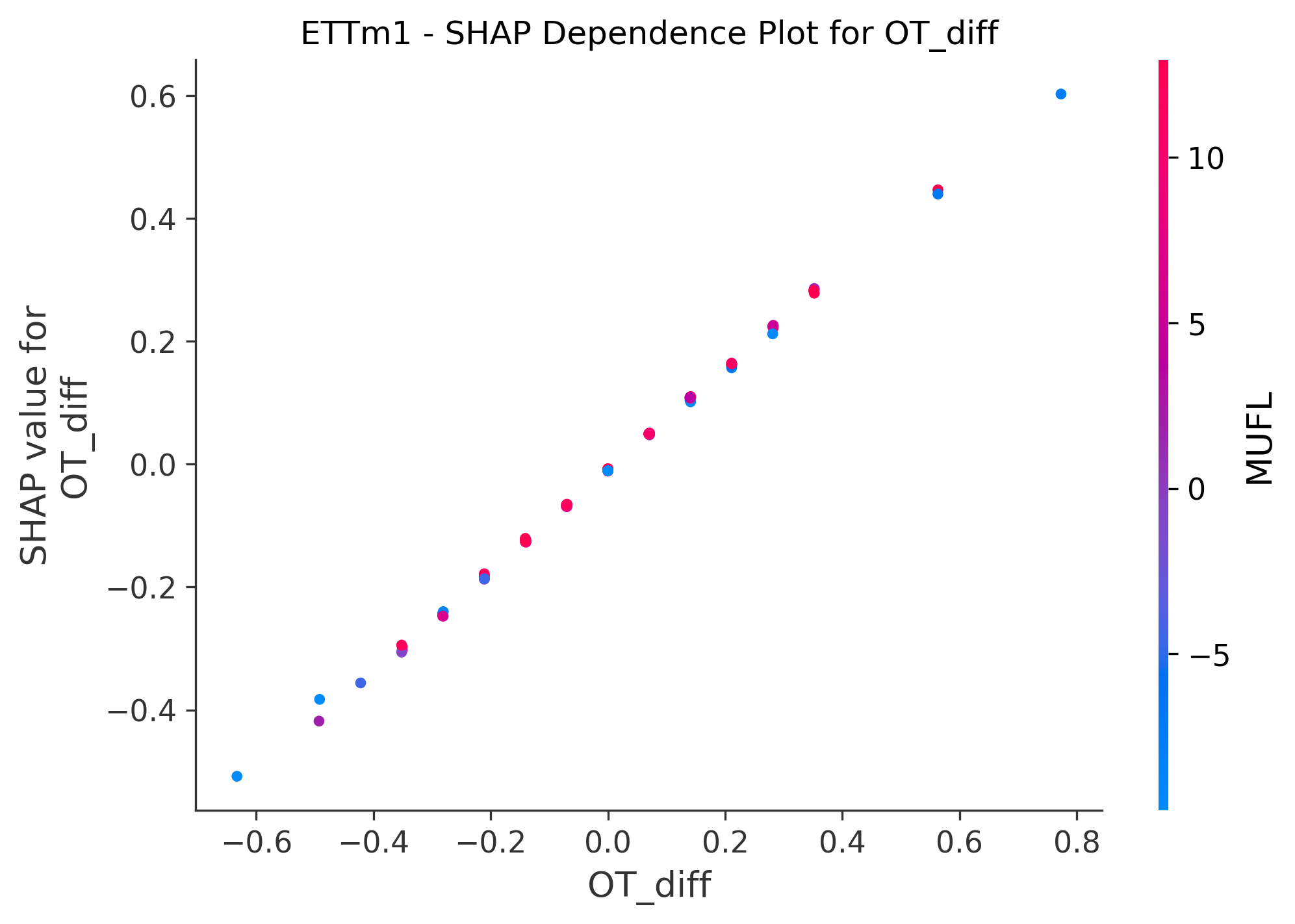}
	\caption{ETTm1 - SHAP Dependence Plot for OT\_diff}
	\label{fig:ettm1_shap_dependence}
\end{figure}

\begin{figure}[htbp]
	\centering
	\includegraphics[width=0.8\textwidth]{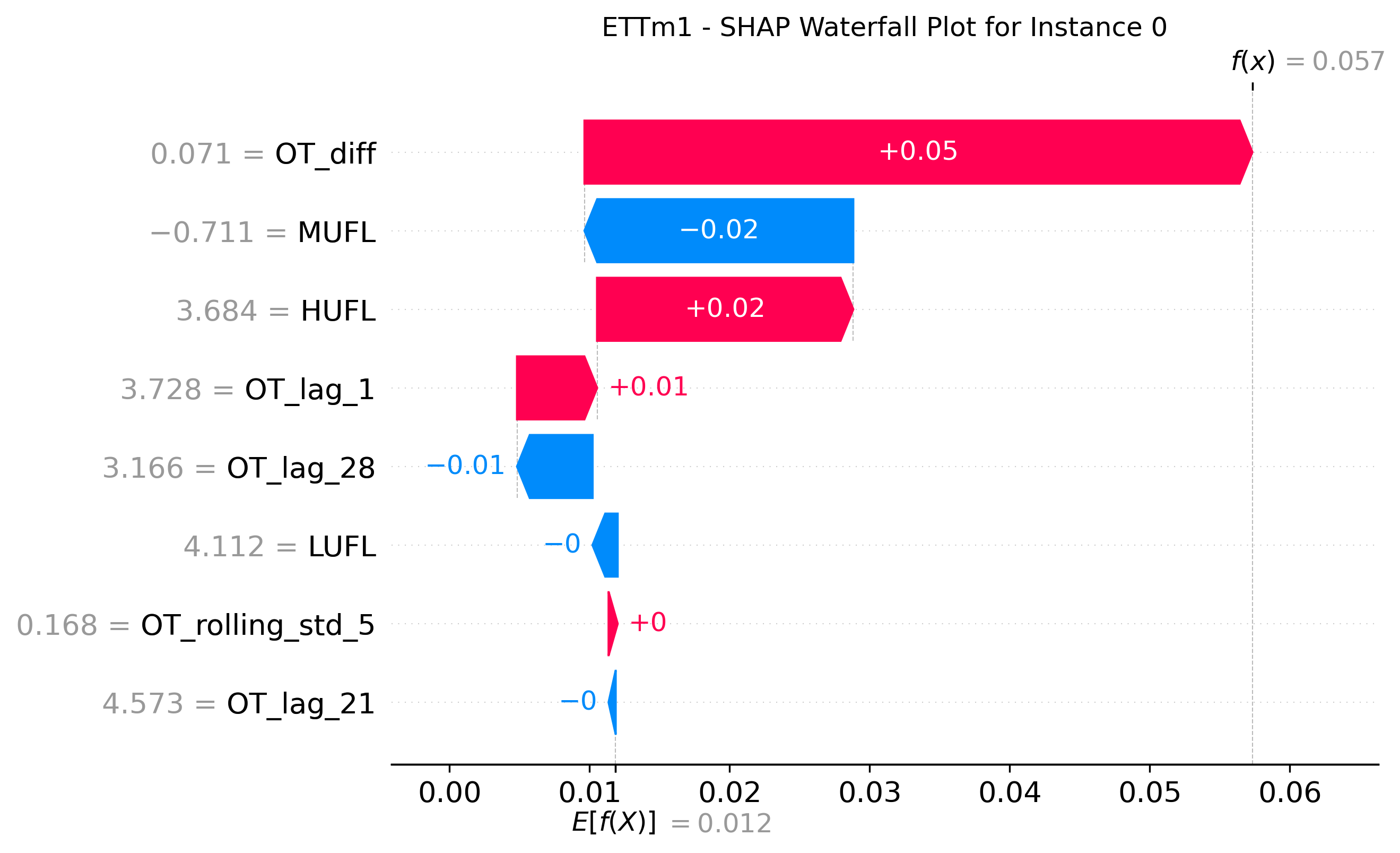}
	\caption{ETTm1 - SHAP Waterfall Plot for Instance 0}
	\label{fig:ettm1_shap_waterfall}
\end{figure}

For the ETTm1 dataset, characterized by minute\_level granularity, the SHAP bar plot (Figure~\ref{fig:ettm1_shap_bar}) similarly highlights OT\_diff as the most significant feature, with HUFL and MUFL as secondary contributors. The summary plot (Figure~\ref{fig:ettm1_shap_summary}) indicates that higher OT\_diff values increase the predicted output, in contrast to a negative influence from MUFL. This relationship is further explored in the dependence plot (Figure~\ref{fig:ettm1_shap_dependence}), which shows a positive trend for OT\_diff, modulated by MUFL interactions. The waterfall plot for Instance 0 (Figure~\ref{fig:ettm1_shap_waterfall}) reveals a positive contribution from OT\_diff (+0.05) to the prediction (\(f(x) = 0.057\)), balanced by a negative effect from MUFL (-0.02) and a positive contribution from HUFL (+0.02). Table~\ref{tab:ettm1_contributions} provides a detailed breakdown, illustrating the interplay of multivariate features at this finer temporal scale.

\begin{table}[htbp]
	\centering
	\caption{Local Feature Contributions for Instance 0 (ETTm1)}
	\label{tab:ettm1_contributions}
	\begin{tabular}{|l|c|c|c|}
		\hline
		Feature & SHAP Value & Feature Value & Contribution Impact \\
		\hline
		OT\_diff & 0.047680 & 0.071000 & 0.003385 \\
		MUFL & -0.019186 & -0.711000 & 0.013641 \\
		HUFL & 0.018289 & 3.684000 & 0.067378 \\
		OT\_lag\_1 & 0.005713 & 3.728000 & 0.021299 \\
		OT\_lag\_28 & -0.005390 & 3.166000 & -0.017064 \\
		LUFL & -0.001790 & 4.112000 & -0.007363 \\
		OT\_rolling\_std\_5 & 0.000642 & 0.167877 & 0.000108 \\
		OT\_lag\_21 & -0.000500 & 4.573000 & -0.002287 \\
		\hline
	\end{tabular}
\end{table}

\subsection{Waste Dataset Analysis}

\begin{figure}[htbp]
	\centering
	\begin{subfigure}{0.49\textwidth}
		\includegraphics[width=\textwidth,height=7cm]{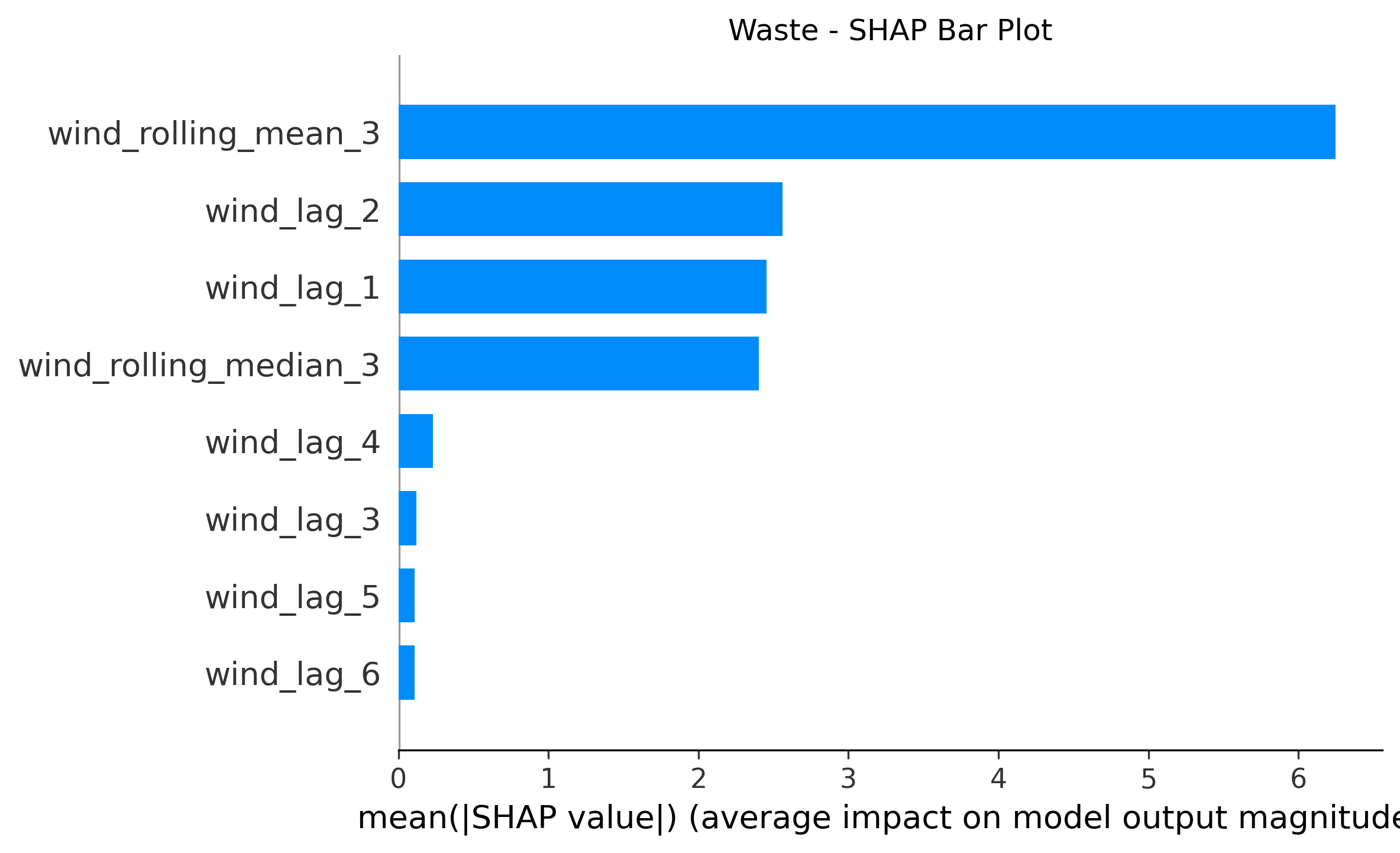}
		\caption{SHAP Bar Plot}
		\label{fig:waste_shap_bar}
	\end{subfigure}
	\hfill
	\begin{subfigure}{0.49\textwidth}
		\includegraphics[width=\textwidth,height=7cm]{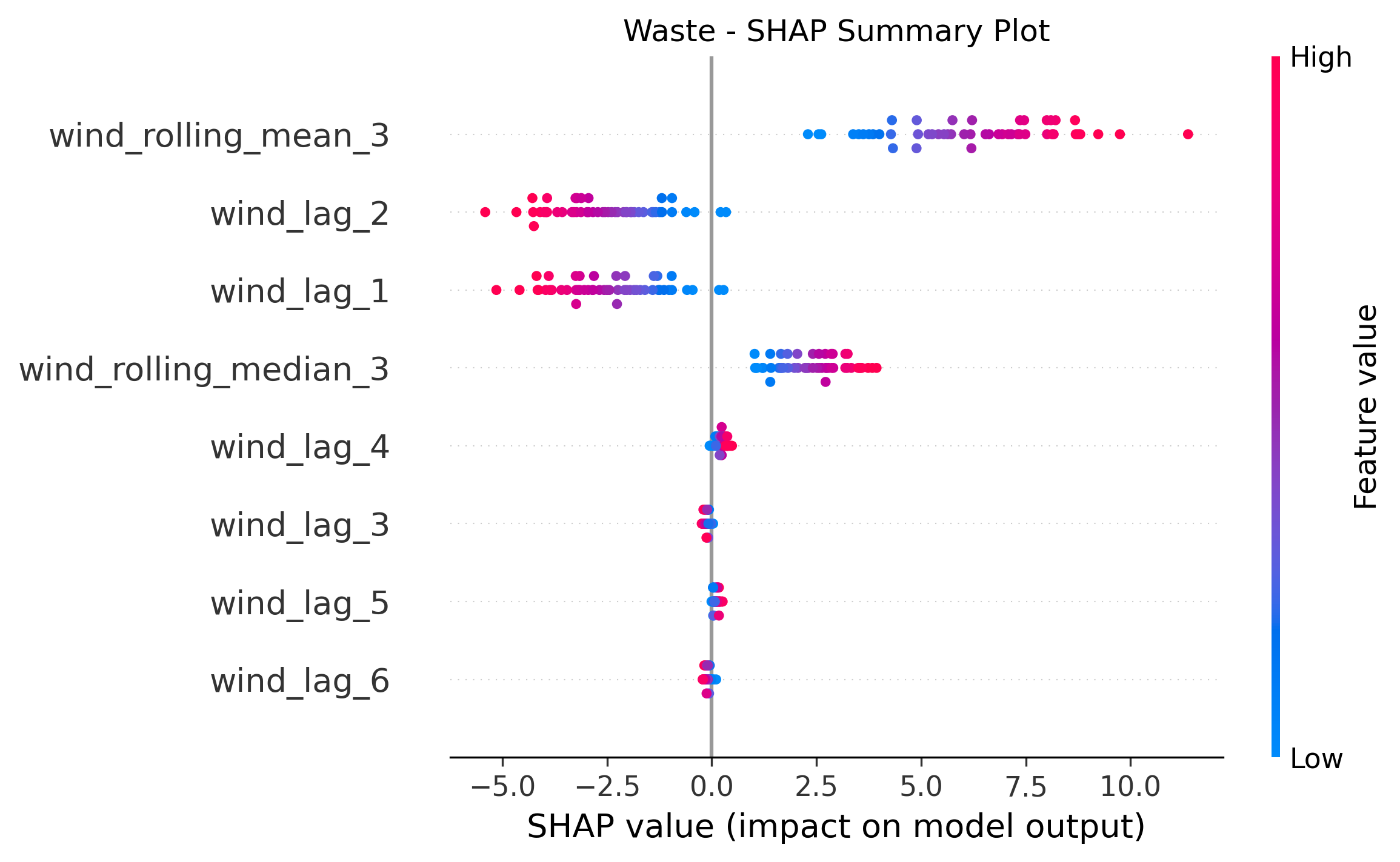}
		\caption{SHAP Summary Plot}
		\label{fig:waste_shap_summary}
	\end{subfigure}
	\caption{Waste - Global SHAP Visualizations}
\end{figure}

\begin{figure}[htbp]
	\centering
	\includegraphics[width=0.8\textwidth]{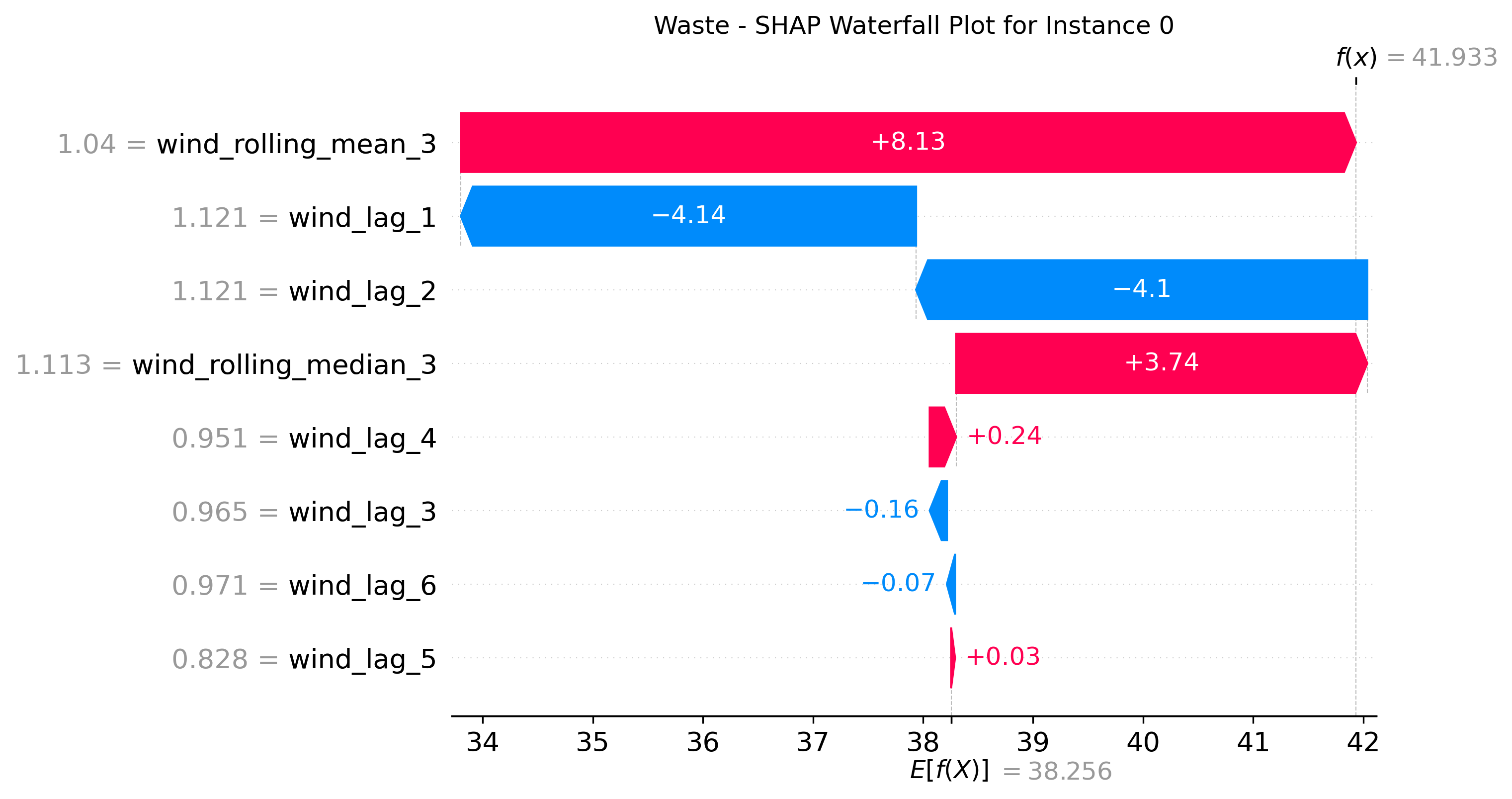}
	\caption{Waste - SHAP Waterfall Plot for Instance 0}
	\label{fig:waste_shap_waterfall}
\end{figure}

The SHAP analysis for the waste dataset, which focuses on wind\_related energy consumption, identifies wind\_rolling\_mean\_3 as the most influential feature, as shown in the bar plot (Figure~\ref{fig:waste_shap_bar}), with an average impact exceeding 5 on the model output magnitude. Other significant features include wind\_lag\_2, wind\_lag\_1, and wind\_rolling\_median\_3, each with impacts around 3-4, while higher-order lags (wind\_lag\_4 to wind\_lag\_6) contribute minimally, with impacts less than 1. The summary plot (Figure~\ref{fig:waste_shap_summary}) indicates that higher values of wind\_rolling\_mean\_3 generally increase the predicted output, with a positive spread of SHAP values, whereas wind\_lag\_1 and wind\_lag\_2 show mixed effects, with negative contributions at higher feature values. At the local level, the waterfall plot for Instance 0 (Figure~\ref{fig:waste_shap_waterfall}) reveals a final prediction of \( f(x) = 41.933 \), starting from an expected value of \( E[f(x)] = 38.256 \). The dominant positive contribution comes from wind\_rolling\_mean\_3 (+8.13), offset by negative contributions from wind\_lag\_1 (-4.14) and wind\_lag\_2 (-4.1), with wind\_rolling\_median\_3 adding +3.74. Minor contributions from wind\_lag\_4 (+0.24) and others adjust the final output.

\subsection{Hydroelectric Dataset Analysis}

\begin{figure}[htbp]
	\centering
	\begin{subfigure}{0.49\textwidth}
		\includegraphics[width=\textwidth,height=7cm]{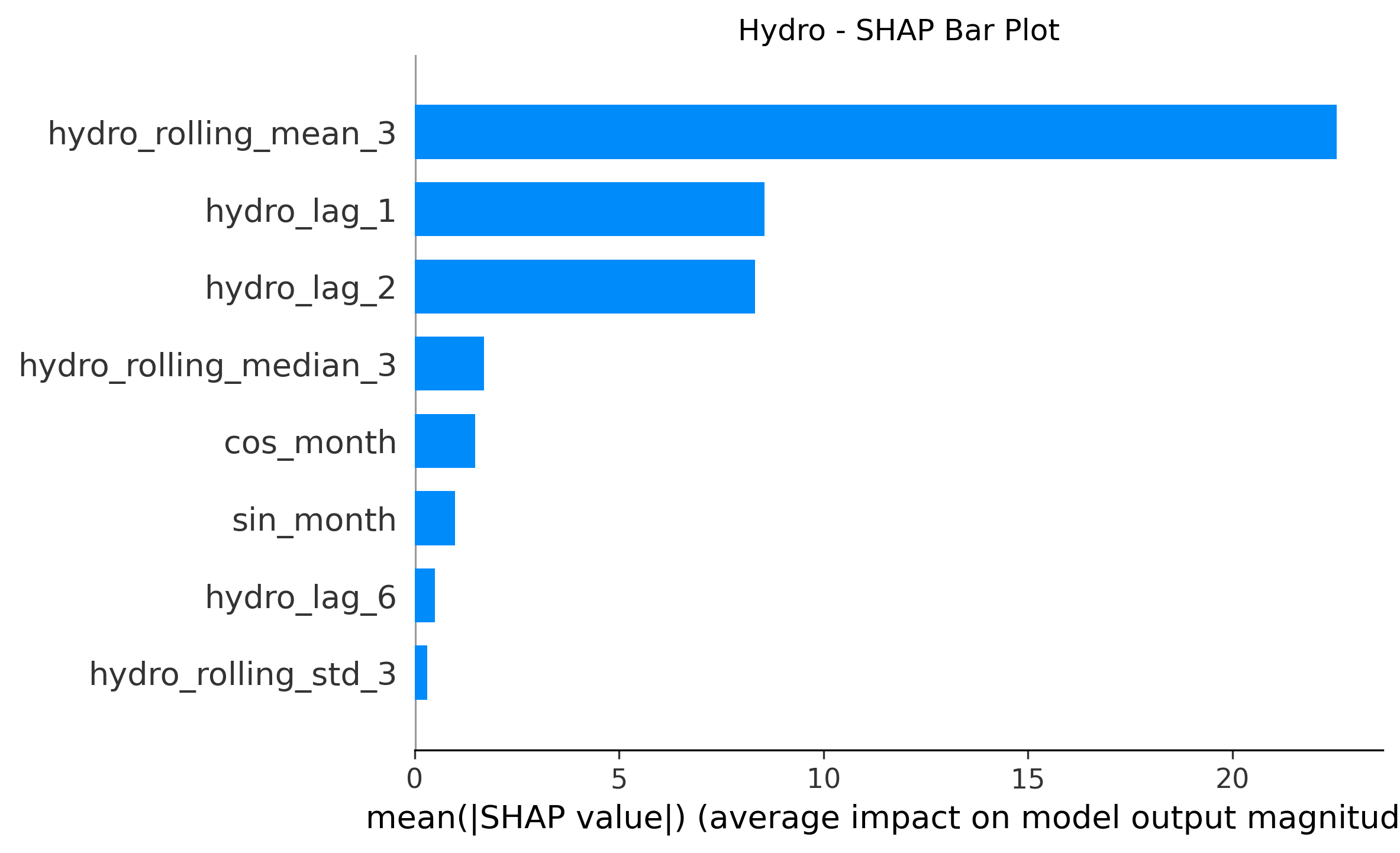}
		\caption{SHAP Bar Plot}
		\label{fig:hydro_shap_bar}
	\end{subfigure}
	\hfill
	\begin{subfigure}{0.49\textwidth}
		\includegraphics[width=\textwidth,height=7cm]{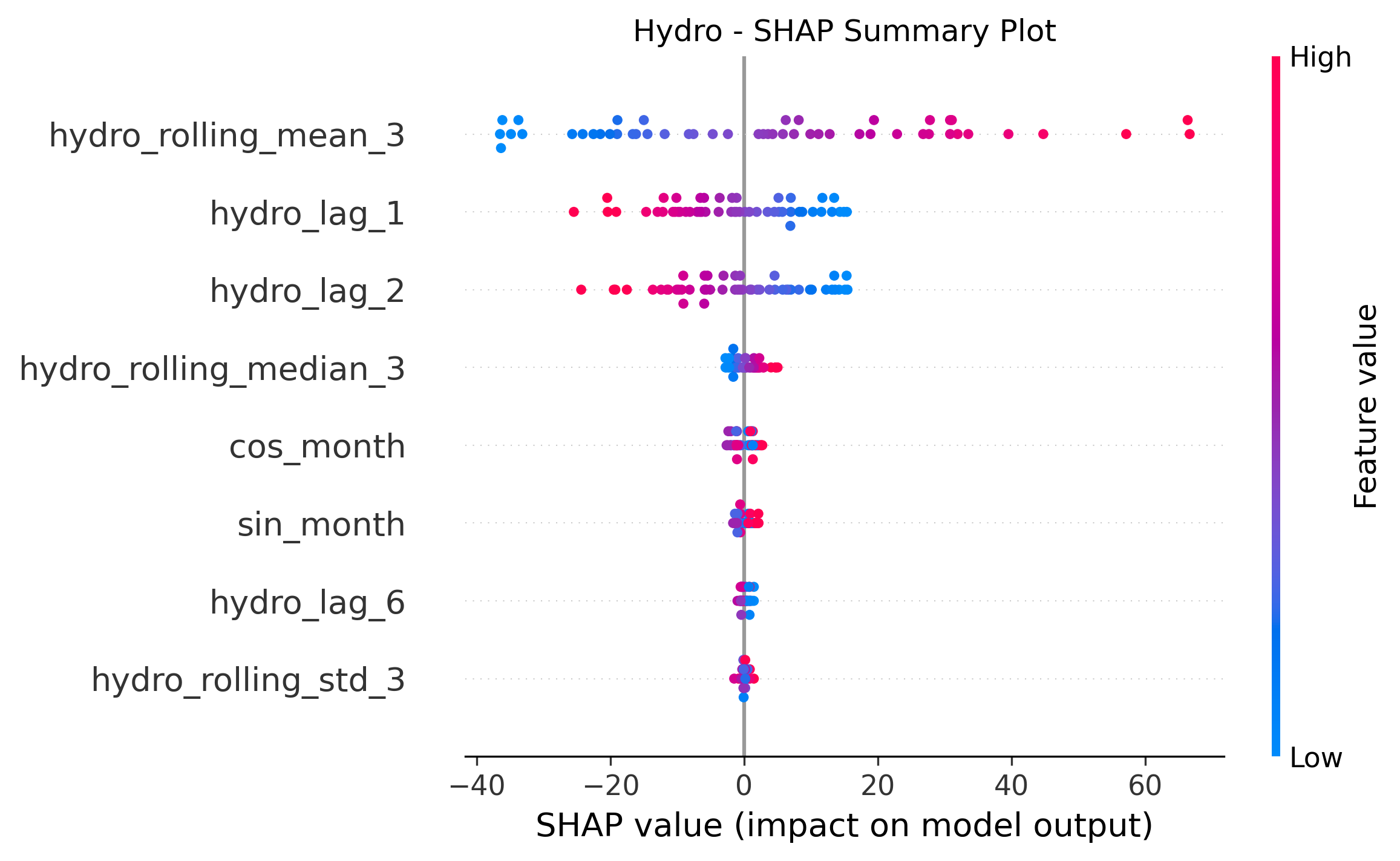}
		\caption{SHAP Summary Plot}
		\label{fig:hydro_shap_summary}
	\end{subfigure}
	\caption{Hydro - Global SHAP Visualizations}
\end{figure}

\begin{figure}[htbp]
	\centering
	\includegraphics[width=0.8\textwidth]{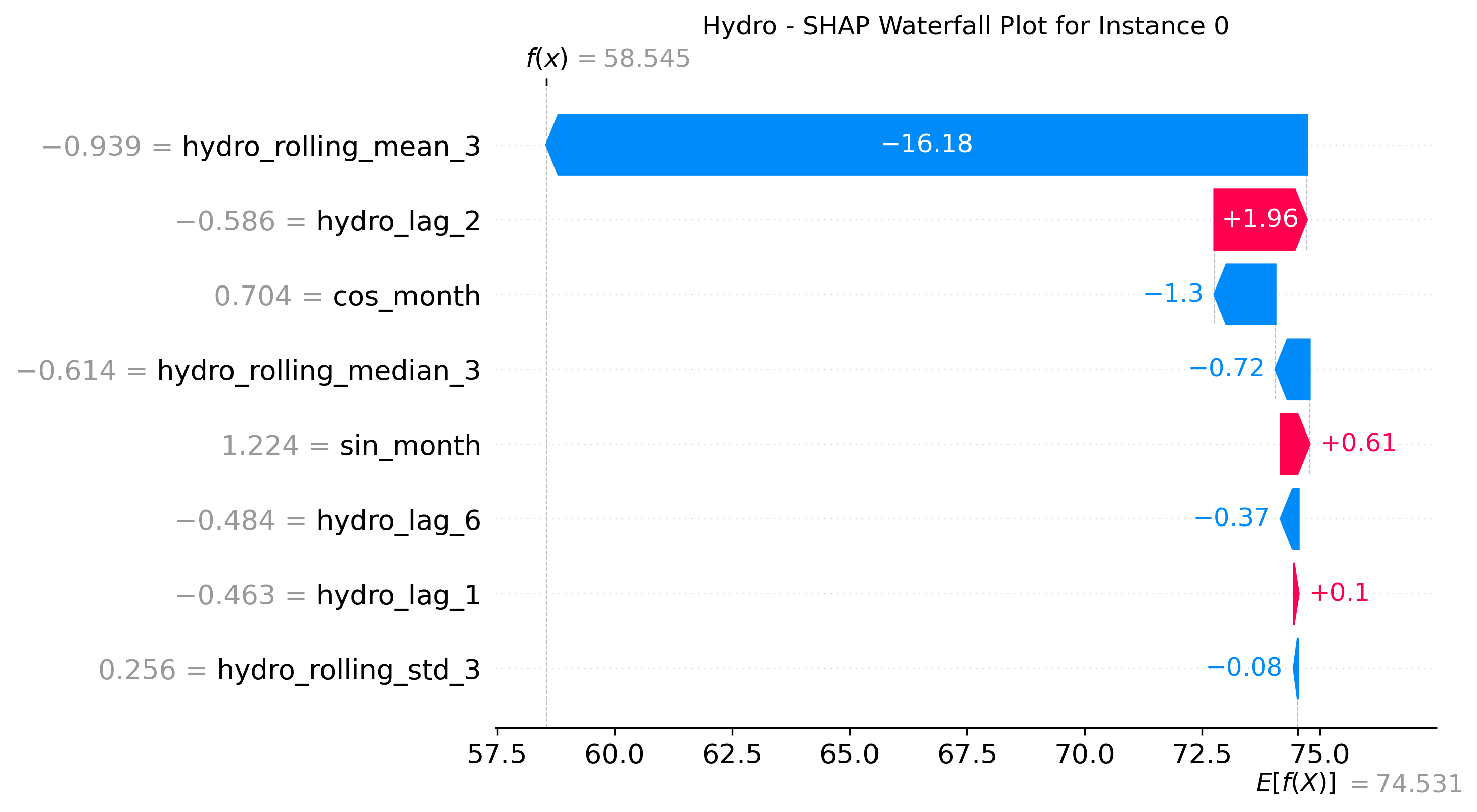}
	\caption{Hydro - SHAP Waterfall Plot for Instance 0}
	\label{fig:hydro_shap_waterfall}
\end{figure}

For the hydroelectric dataset, the SHAP bar plot (Figure~\ref{fig:hydro_shap_bar}) highlights hydro\_rolling\_mean\_3 as the most impactful feature, with an average SHAP value exceeding 15, followed by hydro\_lag\_1 and hydro\_lag\_2, each with impacts around 5-10. Features like hydro\_rolling\_median\_3, cos\_month, sin\_month, hydro\_lag\_6, and hydro\_rolling\_std\_3 have significantly lower impacts, all below 5. The summary plot (Figure~\ref{fig:hydro_shap_summary}) reveals that higher hydro\_rolling\_mean\_3 values negatively affect the output (SHAP values extending to -40), while hydro\_lag\_1 and hydro\_lag\_2 exhibit positive contributions at higher values. The waterfall plot for Instance 0 (Figure~\ref{fig:hydro_shap_waterfall}) shows a prediction of \( f(x) = 58.545 \), starting from \( E[f(x)] = 74.531 \). The largest negative contribution is from hydro\_rolling\_mean\_3 (-16.18), partially offset by positive contributions from hydro\_lag\_2 (+1.96) and sin\_month (+0.61), with cos\_month (-1.3) and hydro\_rolling\_median\_3 (-0.72) further modulating the outcome.

\subsection{Solar Dataset Analysis}

\begin{figure}[htbp]
	\centering
	\begin{subfigure}{0.49\textwidth}
		\includegraphics[width=\textwidth,height=7cm]{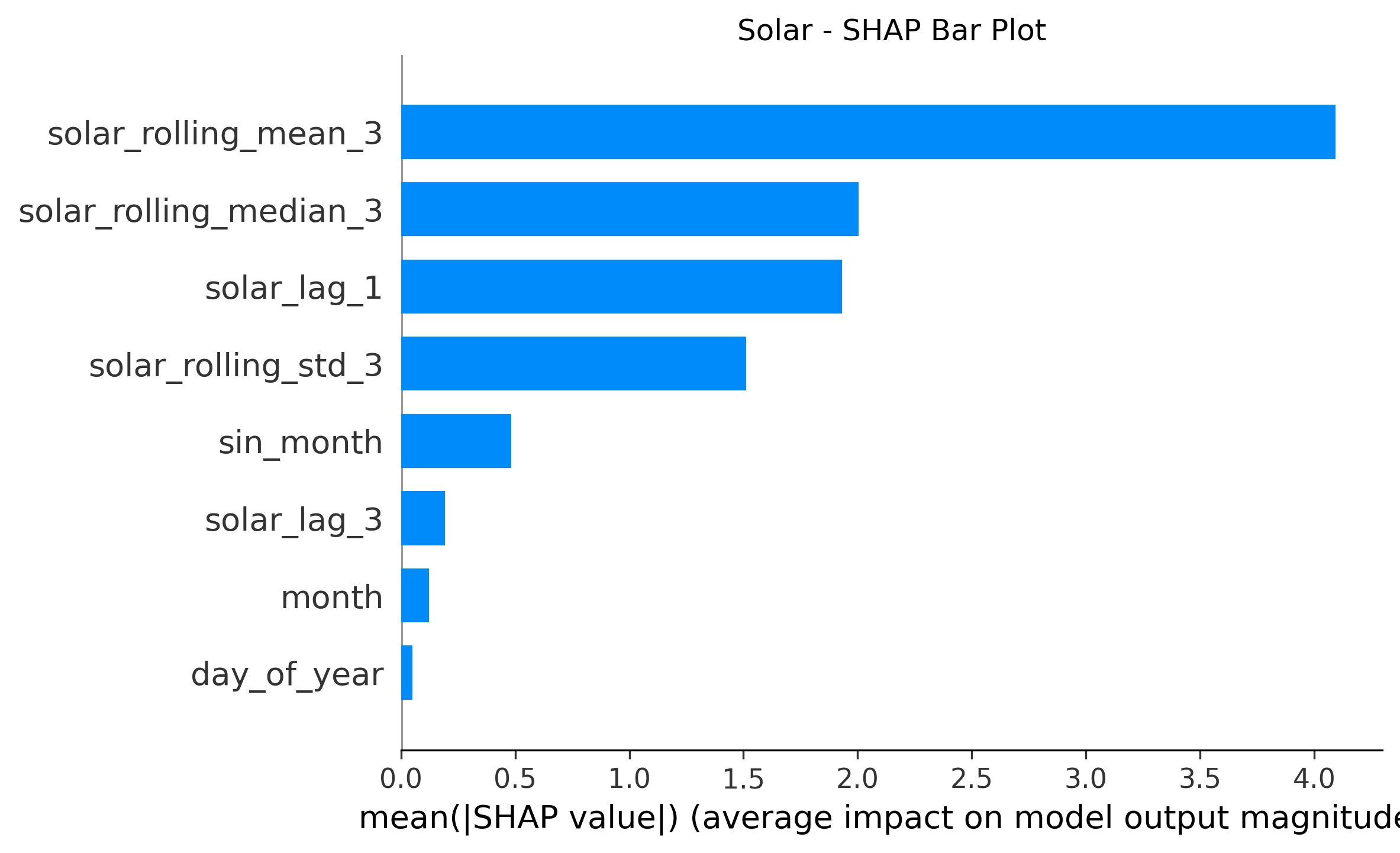}
		\caption{SHAP Bar Plot}
		\label{fig:solar_shap_bar}
	\end{subfigure}
	\hfill
	\begin{subfigure}{0.49\textwidth}
		\includegraphics[width=\textwidth,height=7cm]{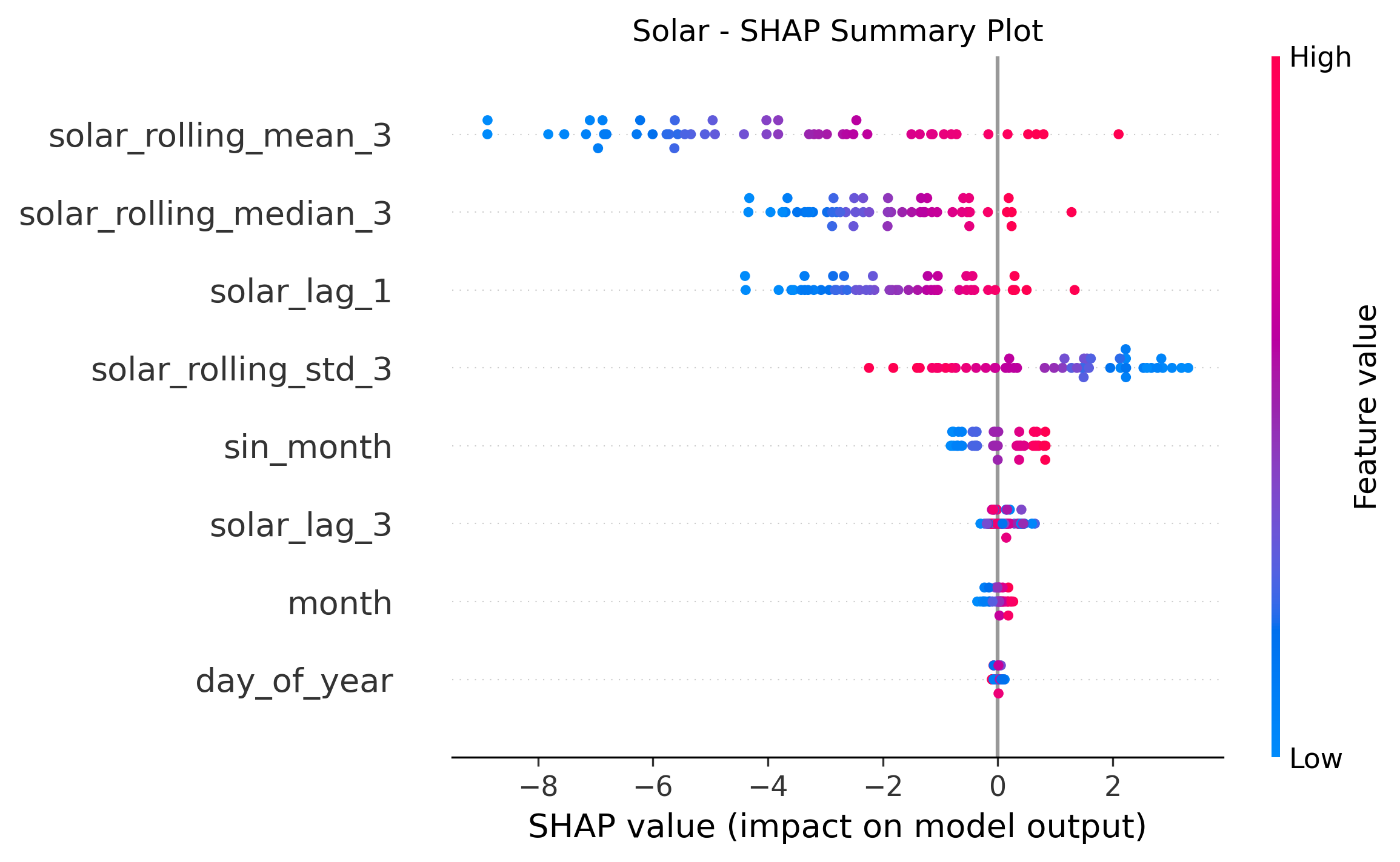}
		\caption{SHAP Summary Plot}
		\label{fig:solar_shap_summary}
	\end{subfigure}
	\caption{Solar - Global SHAP Visualizations}
\end{figure}

\begin{figure}[htbp]
	\centering
	\includegraphics[width=0.8\textwidth]{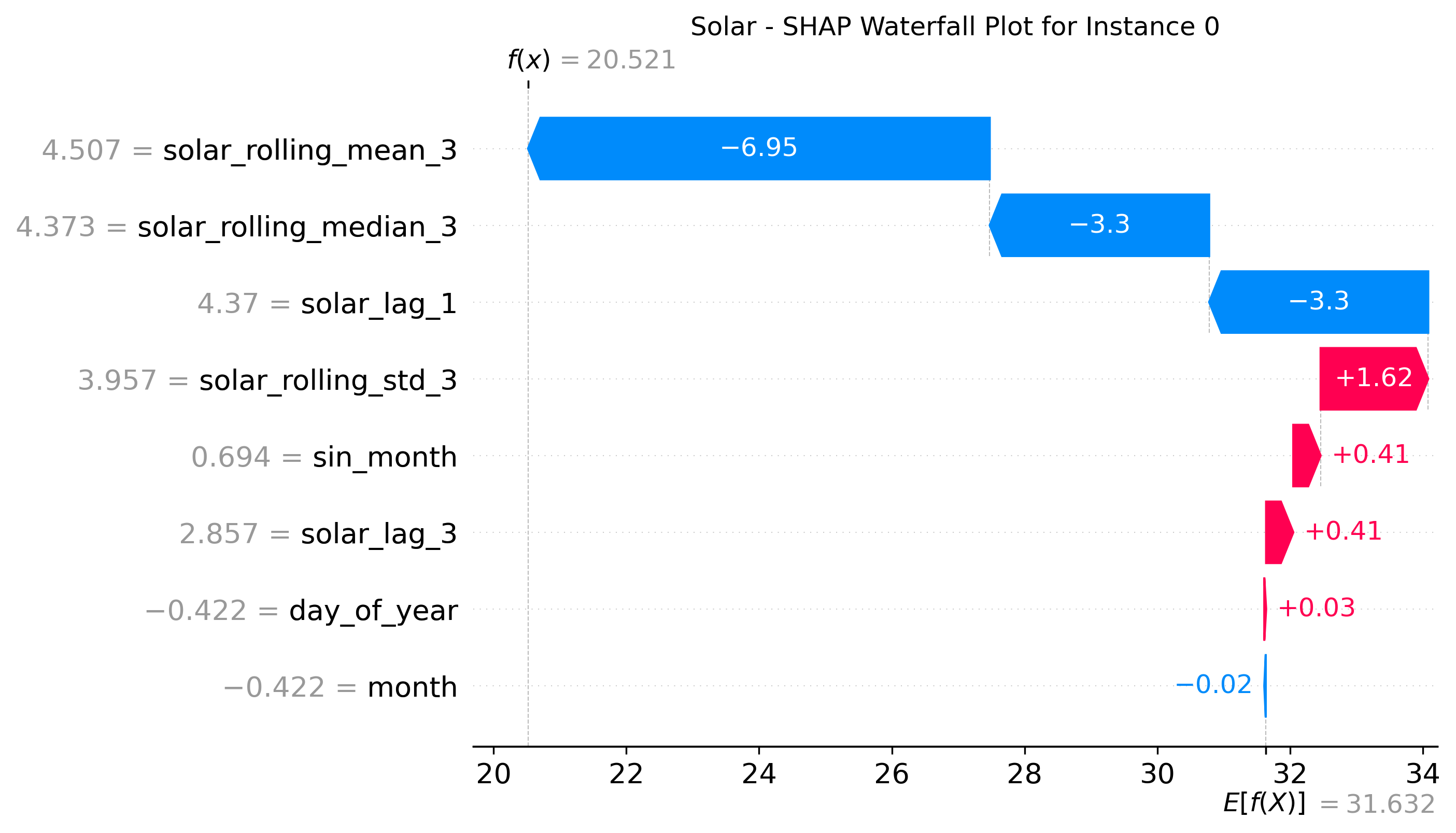}
	\caption{Solar - SHAP Waterfall Plot for Instance 0}
	\label{fig:solar_shap_waterfall}
\end{figure}

The SHAP analysis for the solar dataset identifies solar\_rolling\_mean\_3 as the most significant feature in the bar plot (Figure~\ref{fig:solar_shap_bar}), with an average impact around 3.5, followed closely by solar\_rolling\_median\_3, solar\_lag\_1, and solar\_rolling\_std\_3, each with impacts around 2-3. Seasonal features like sin\_month and temporal features such as month and day\_of\_year have minimal influence, with impacts less than 1. The summary plot (Figure~\ref{fig:solar_shap_summary}) shows that higher solar\_rolling\_mean\_3 values increase the output (positive SHAP spread), while solar\_rolling\_std\_3 exhibits a negative effect at higher values. The waterfall plot for Instance 0 (Figure~\ref{fig:solar_shap_waterfall}) yields a prediction of \( f(x) = 20.521 \), starting from \( E[f(x)] = 31.632 \). The largest negative contribution comes from solar\_rolling\_mean\_3 (-6.95), with additional reductions from solar\_rolling\_median\_3 (-3.3) and solar\_lag\_1 (-3.3), offset by positive contributions from solar\_rolling\_std\_3 (+1.62) and sin\_month (+0.41).

\subsection{Comparative Insights and Implications}

The SHAP-based interpretability analysis across the ETTh1, ETTm1, waste, hydro, and solar datasets reveals both commonalities and distinct patterns in feature importance. In the ETTh1 and ETTm1 datasets, OT\_diff consistently emerges as the dominant feature, yet its impact varies by temporal resolution—negative at the hourly level (ETTh1) and positive at the minute\_level (ETTm1)—highlighting the sensitivity of differential features to time scales. Multivariate features like HUFL and MUFL play significant roles in both datasets, with their effects modulated by interactions with OT\_diff. For the waste dataset, wind\_rolling\_mean\_3 dominates with a strong positive influence, reflecting the importance of short-term statistical trends in wind energy forecasting. The hydroelectric dataset shows a pronounced negative impact from hydro\_rolling\_mean\_3, suggesting that averaging over three periods may over-smooth critical variations, with seasonal features like cos\_month and sin\_month providing moderate adjustments. In the solar dataset, solar\_rolling\_mean\_3 also dominates but with a negative contribution, indicating a potential over-reliance on mean trends that may not capture solar-specific variability, mitigated by solar\_rolling\_std\_3 and seasonal effects.

These findings underscore the critical role of temporal lag and statistical features (e.g., rolling means and medians) across all datasets, with seasonal components gaining prominence in hydro and solar due to their periodic nature. The varying directions of influence (positive in waste, negative in hydro and solar) suggest dataset-specific dynamics that warrant tailored feature engineering. Future work can leverage these insights to optimize the selection and transformation of lag and rolling features, potentially incorporating interaction terms or alternative statistical measures to enhance predictive accuracy and interpretability in energy forecasting systems.

\section{Conclusion, Limitations, and Future Directions}

This research propose a neural architecture that fuses Neural Ordinary Differential Equations using  Runge-Kutta method, graph attention mechanisms, and Daubechies wavelet transforms to enhance time series forecasting in multi-sensor energy systems. By integrating continuous-time dynamics, relational inter-feature dependencies, and multi-scale temporal patterns, the model adeptly processes heterogeneous data from diverse sources. Evaluated on the ETT datasets (ETTh1, ETTh2, ETTm1, ETTm2) and renewable energy datasets (Waste, Solar, Hydro), it demonstrates superior performance, achieving an MSE of 0.0135 and MAE of 0.0843 on ETTh1 in non-windowed settings—outpacing baselines like N-BEATS, N-HiTS, and TCN—and maintaining long-term accuracy (MAE of 0.0513 at 720 steps). On windowed data, it always gives the best RMSE and MAE scores for each horizon, showing that it is adaptable. For renewable energy, it excels on Waste (R\textsuperscript{2} of 0.9822) and Hydro (R\textsuperscript{2} of 0.9887), though it slightly lags an Attention-based model on Solar (R\textsuperscript{2} of 0.8716 vs. 0.8886) due to challenges with irregular solar patterns. SHAP analysis makes the combination of local wavelet features, global Neural ODE dynamics and graph-based insights better at predicting energy and temperature, as well as easier to understand.

Even with its advantages, there are still some limitations. Since feature engineering is mostly done manually, the approach is not easily scalable to many different multi-sensor domains, so automated solutions are needed. The fixed temporal feature set may limit adaptation to dynamically evolving pat- terns critical for real-time multi-source forecasting. Also, since Neural ODEs, graph attention and multi-resolution transforms are computationally complex, deploying them to resource-limited edge devices is hard. The error metrics computation on Solar dataset shows that it is not easy for the model to combine the unique and irregular patterns from different energy sources, so special fusion approaches are needed. Evaluations are thorough but restricted to certain datasets which could make it hard to use them for many multi-sensor scenarios. The adaptive frequency layer’s way of combining information is not easily interpretable which may make it less suitable for cases where full transparency is required. Furthermore, the graph attention mechanism’s reliance on Pearson correlations for dynamic graph construction may fail to capture nonlinear relationships, reducing the interpretability of feature interactions in black-box attention scenarios.

Future work should enhance this fusion framework by developing automated feature extraction to improve scalability, optimizing computational efficiency for edge deployment, and incorporating real-time adaptation for dynamic multi-sensor data. Expanding evaluations to diverse, large-scale datasets and refining fusion techniques for irregular patterns (e.g., solar energy) could bolster robustness. Advancing interpretability through explainable AI methods would further clarify the fusion of multi-source inputs, supporting critical applications. This architecture lays a strong foundation for multi-sensor time series forecasting in energy systems, demonstrating the power of information fusion, while identifying key areas—automation, efficiency, and interpretability—for future exploration to maximize its impact across broader domains.

\section*{Declaration of competing interest}
The author declares that he has no known competing financial interests or personal relationships that could have appeared to influence the work reported in this paper.

\section*{Data availability}
The dataset used in this study has been cited within the paper.

\section*{CRediT authorship contribution statement}
\textbf{Usman Gani Joy}: Writing – original draft, Writing – Review \& editing.

\section*{Funding Source}
This research did not receive any specific grant from funding agencies in the public, commercial, or not-for-profit sectors.

\section*{Declaration of generative AI and AI-assisted technologies in the writing process}
Statement: During the preparation of this work, large language models (LLMs) were utilized for paraphrasing and proofreading purposes. The content was subsequently reviewed and edited as necessary, with full responsibility taken for the published article.



\bibliographystyle{elsarticle-num} 
\bibliography{sn-bibliography}




\end{document}